\documentclass[11pt]{article}

\usepackage[preprint]{acl}

\usepackage{hyperref}
\usepackage{fontawesome5}
\usepackage{times}
\usepackage{latexsym}

\usepackage[table,dvipsnames]{xcolor}
\usepackage{graphicx}

\usepackage{bm}
\usepackage{booktabs}
\usepackage{amssymb}

\usepackage{booktabs}

\usepackage{array}
\usepackage{makecell}
\usepackage{multirow}

  \newcommand{\na}{--}
  \newcommand{\best}[1]{\textbf{#1}}
  \newcommand{\second}[1]{\underline{#1}}

\newcommand{\gain}[1]{\textcolor{ForestGreen}{#1}}
\newcommand{\loss}[1]{\textcolor{BrickRed}{#1}}
\newcommand{\neutral}[1]{\textcolor{gray}{#1}}

\definecolor{mygreen}{HTML}{1B9E77}

\usepackage[T1]{fontenc}

\usepackage[utf8]{inputenc}

\usepackage{longtable}
\usepackage{listings}
\usepackage{microtype}

\usepackage{inconsolata}

\usepackage{graphicx}

\title{\textsc{DocOps}: A Verifiable Benchmark for Autonomous Agents in \\ Complex Document Operations}

\author{
  \textbf{Jiazhen Jiang}\textsuperscript{1,2} \quad
  \textbf{Boxi Cao}\textsuperscript{1} \quad
  \textbf{Lingyong Yan}\textsuperscript{3} \quad
  \textbf{Yaojie Lu}\textsuperscript{1} \quad
  \textbf{Hongyu Lin}\textsuperscript{1}
  \\
  \textbf{Shuaiqiang Wang}\textsuperscript{3} \quad
  \textbf{Dawei Yin}\textsuperscript{3} \quad
  \textbf{Xianpei Han}\textsuperscript{1} \quad
  \textbf{Le Sun}\textsuperscript{1}
  \\
  \makecell{
    \textsuperscript{1}Chinese Information Processing Laboratory \\
    Institute of Software, Chinese Academy of Sciences
  }
  \\
  \textsuperscript{2}University of Chinese Academy of Sciences
  \quad
  \textsuperscript{3}Baidu Inc.
  \\
  \texttt{\{jiangjiazhen2025, caoboxi\}@iscas.ac.cn}
  \\
  \href{https://docopsbench.github.io}
  {\textcolor{MidnightBlue}{\faLink}\hspace{0.25em}
  \texttt{https://docopsbench.github.io}}
}

\begin{document}

\maketitle
\begin{abstract}
As autonomous agents rapidly evolve, their ability to reliably manipulate ubiquitous digital documents has become critical for enabling general-purpose AI assistants and automating complex workspace workflows. 
In this paper, we introduce \textbf{DocOps}, a deterministically verifiable evaluation framework underpinned by a hierarchical taxonomy that deconstructs document operations inspired by real-world practices into atomic dimensions and escalating workflow complexities.
Based on DocOps, we systematically evaluate representative closed- and open-source models across various agentic harnesses, revealing that even the most advanced frontier configurations still exhibit profound limitations when handling highly coupled, long-range tasks. Furthermore, a fine-grained analysis of existing agents' manipulation behaviors uncovers 3 key failure modes: long-term state tracking collapse, shallow semantic verification, and destructive editing of structural metadata. Ultimately, our work exposes the capability boundaries of agents in maintaining global document consistency, shedding light on the future design of robust, non-destructive agents for complex digital ecosystems.\footnote{Both the code and dataset are publicly available: \url{https://github.com/icip-cas/DocOps}.}

\end{abstract}

\begin{figure*}[t]
\centering
\setlength{\abovecaptionskip}{0.1cm}
\setlength{\belowcaptionskip}{-0.6cm}
\includegraphics[width=0.9\textwidth]{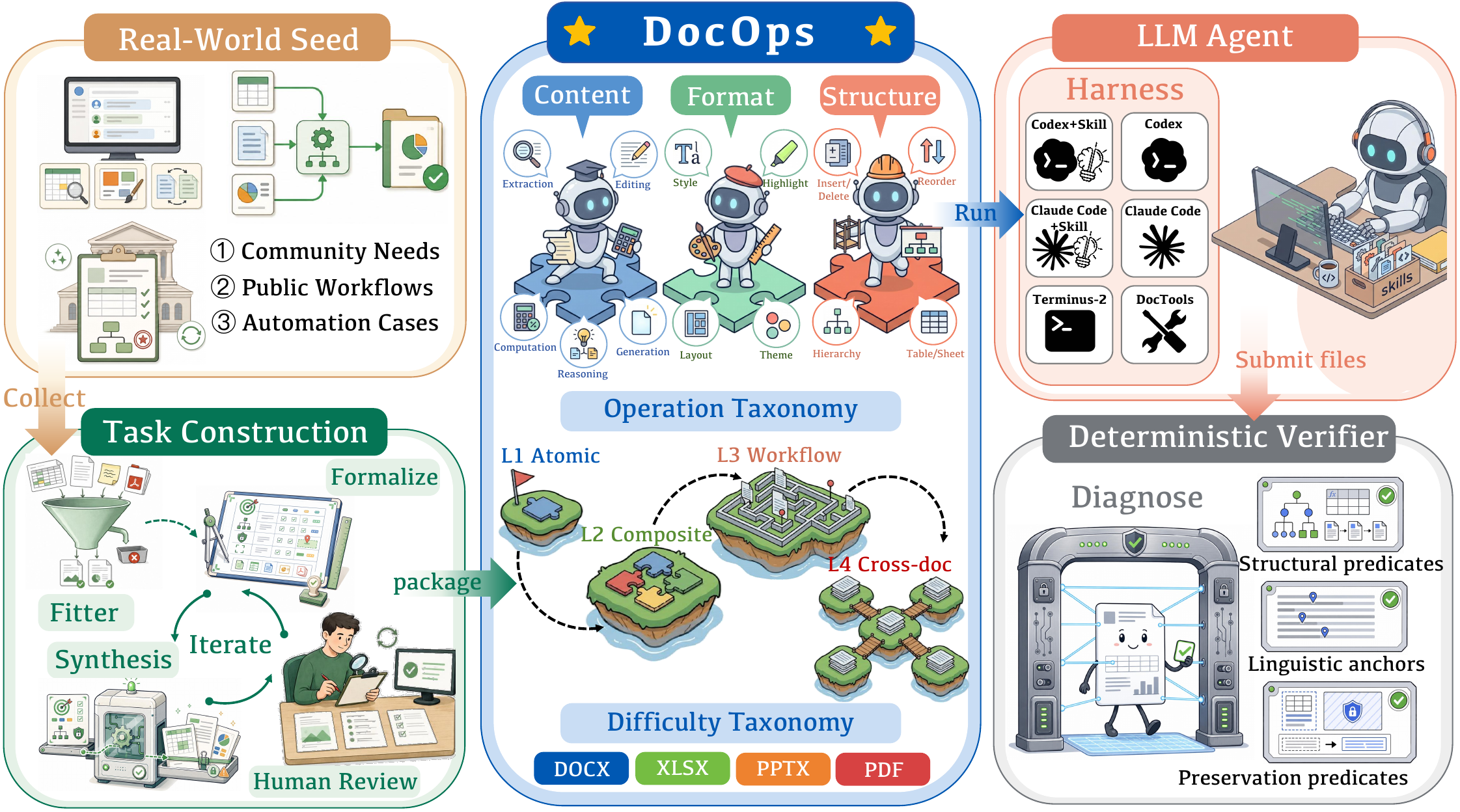}
\caption{
The overview framework of DocOps. 
}
\label{fig:overview}
\end{figure*}

\section{Introduction}

Autonomous agents powered by large language models (LLMs) are rapidly evolving from passive conversationalists to active participants in complex digital workspaces~\citep{10.5555/3692070.3692532, koh-etal-2024-visualwebarena, NEURIPS2024_5d413e48}. A central pillar of human productivity involves the creation, modification, and management of ubiquitous digital documents, such as spreadsheets, presentations, word-processed files, and PDFs~\citep{NEURIPS2023_0ff30c4b, wang2024officebenchbenchmarkinglanguageagents}. As we push toward general-purpose AI assistants, the ability to autonomously navigate and reliably manipulate these highly structured digital artifacts represents a critical frontier in human-computer interaction~\citep{ge2023llmosagentsapps, trivedi-etal-2024-appworld}.


Despite rapid advancements in agentic capabilities, our quantitative understanding of how models interact with real-world documents remains fundamentally constrained by 2 paradigms~\citep{Wang_2024}. 
On the one hand, static frameworks (e.g., DocBench~\citep{zou2024docbenchbenchmarkevaluatingllmbased}, DUDE~\citep{DBLP:conf/iccv/LandeghemPTJBBC23}) reduce documents to read-only knowledge repositories, benchmarking only information extraction or question-answering over fixed text. 
On the other end, workflow-oriented evaluations (e.g., OfficeBench~\citep{wang2024officebenchbenchmarkinglanguageagents}, OdysseyBench~\citep{wang2025odysseybenchevaluatingllmagents}) focus primarily on the orchestration of software navigation, treating the document merely as a transient, passive data payload passed between different applications. 
Crucially, neither approach treats the document as a first-class computational object, an intricate digital artifact that an agent must actively and continuously manipulate throughout a complex workflow. In real-world productivity scenarios, agents are required to directly synthesize and alter diverse document contents and formats, ensuring that the modified files remain structurally valid and functionally intact. 
Consequently, it remains an open research question \textit{whether current agents can reliably execute end-to-end document-centric tasks while maintaining global document state consistency, and achieving user objectives without introducing destructive modifications.}

To this end, we introduce \textbf{DocOps}, a rigorously verifiable evaluation framework designed to systematically assess LLM agents on complex end-to-end document manipulation. 
A primary contribution of this work is the formulation of a principled taxonomy that systematically deconstructs the operational black box of document manipulation. 
As illustrated in Figure~\ref{fig:overview}, our taxonomy maps the operational space across two orthogonal axes: atomic capabilities and workflow depth. 
First, we isolate document interaction into 3 interdependent dimensions: content, format, and structure. Second, we establish a four-tiered complexity gradient. While L1 and L2 isolate single and multi-dimensional atomic edits, L3 and L4 escalate to long-horizons, cross-document workflows inspired by real-world human interaction scenarios. 
This hierarchical taxonomy serves not merely as a categorization scheme, but as a fine-grained agentic diagnostic framework, enabling researchers to localize whether an agent's failure stems from localized structural disruption or a collapse in long-term state tracking.

Crucially, prior benchmarks often rely on indirect validation targets and coarse task-level success signals, which makes it difficult to fully assess whether agents can produce valid and non-destructively edited office artifacts~\citep{ofengenden2025pptarenabenchmarkagenticpowerpoint, wang2024officebenchbenchmarkinglanguageagents}.
Therefore, to ensure evaluation fidelity within this taxonomy, DocOps introduces a deterministic validator for each task that directly targets the native document.
Specifically, DocOps directly inspects the output files through native document libraries. 
Each verifier consists of 3 types of predicates, which can not only check whether the user-requested editing operations have been correctly completed, but also verify that the associated hidden structural invariants are preserved. This deterministic artifact-level design is intended to reduce overly lenient evaluation in document tasks, and further reveals native document failure modes that cannot be captured by relying solely on surface-level validation, task-level success determination, or automated evaluators.

By systematically evaluating a diverse spectrum of leading closed- and open-source models across various agentic harnesses, we uncover the performance limitations and critical bottlenecks of contemporary agent systems. We show that even the most capable frontier configuration (GPT-5.5~\citep{openai2026gpt55} with Codex and skills) achieves only an overall success rate of 0.671, and its performance rapidly drops on L3 and L4 tasks, demonstrating that current agent systems still struggle with workflow-level document manipulation. 
Further analysis reveals that workflow difficulty increases when operations are tightly coupled. For instance, Excel workflows, which demand strict maintenance of formula references and data validation boundaries, see success rates plummet to near zero in complex long-range tasks, whereas tasks in low-coupling formats like PDFs remain relatively robust.

\newcommand{\cmark}{\ensuremath{\checkmark}}
\newcommand{\xmark}{\ensuremath{\times}}
\newcommand{\bcmark}{\ensuremath{\boldsymbol{\checkmark}}}

\begin{table*}[t]
\centering
\scriptsize
\setlength{\tabcolsep}{4.0pt}
\renewcommand{\arraystretch}{1.12}
\begin{tabular}{@{}llccccc@{}}
\toprule
\multicolumn{1}{c}{\multirow{2}{*}{\textbf{Benchmark}}}
& \multicolumn{1}{c}{\multirow{2}{*}{\textbf{Formats}}}
& \textbf{Native-file}
& \textbf{Cross-doc}
& \textbf{Deterministic}
& \textbf{Explicit}
& \textbf{Harness/skill} \\
\multicolumn{1}{c}{}
& \multicolumn{1}{c}{}
& \textbf{edit}
& \textbf{workflow}
& \textbf{target verifier}
& \textbf{preservation}
& \textbf{analysis} \\
\midrule
DocBench~\citep{zou2024docbenchbenchmarkevaluatingllmbased}
& PDF
& \xmark & \xmark & \xmark & \xmark & \xmark \\

SpreadsheetBench~\citep{NEURIPS2024_ac840df2}
& XLSX
& \cmark & \xmark & \xmark & \xmark & \xmark \\

PPTArena~\citep{ofengenden2025pptarenabenchmarkagenticpowerpoint}
& PPTX
& \cmark & \xmark & \xmark & \xmark & \xmark \\

OfficeBench~\citep{wang2024officebenchbenchmarkinglanguageagents}
& DOCX, XLSX, PDF + apps
& \cmark & \cmark & \cmark & \xmark & \xmark \\

OdysseyBench~\citep{wang2025odysseybenchevaluatingllmagents}
& DOCX, XLSX, PDF + apps
& \cmark & \cmark & \cmark & \xmark & \xmark \\

DELEGATE-52~\citep{laban2026llmscorruptdocumentsdelegate}
& Markdown, CSV, TXT, TeX, etc.
& \cmark & \cmark & \xmark & \cmark & \xmark \\

FINCH~\citep{dong2026finchbenchmarkingfinance}
& XLSX + PDF/DOCX inputs
& \cmark & \cmark & \xmark & \xmark & \xmark \\
\textbf{DocOps (ours)}
& \textbf{XLSX, DOCX, PPTX, PDF}
& \bcmark & \bcmark & \bcmark & \bcmark & \bcmark \\
\bottomrule
\end{tabular}
\caption{Comparison with representative document-understanding, document-editing, and office-workflow benchmarks. A property is marked as present only when it constitutes an explicit benchmark-level evaluation target. ``Deterministic target verifier'' denotes direct programmatic verification of the requested final document state, rather than round-trip reconstruction or LLM-as-a-judge evaluation.}
\label{tab:related_benchmark_comparison}
\end{table*}

Moreover, we identify 3 pervasive failure modes behind agent failures on complex document tasks: 1) \emph{Long-term state tracking failure}, where agents complete local edits but lose global document state. 2) \emph{Shallow semantic verification}, where agents accept surface-plausible outputs without checking underlying computational or structural semantics. 3) \emph{Destructive editing}, where models persistently downgrade complex object trees into flat text or irreparably break formulas, tables or structural metadata. Our quantitative analysis further shows that these modes account for a majority of failed runs.
Finally, we demonstrate that the agent's harness profoundly dictates performance: open-ended programmable coding environments with file-system feedback vastly outperform static RPC-style tool calling. Furthermore, while explicit procedural guidance (e.g., skills) acts as a strong catalyst for mid-tier open-source models, it offers marginal utility for frontier models capable of zero-shot orchestration. 
Together, these findings redefine the trajectory for document-centric AI, shifting the focus from isolated tool invocation to the development of state-aware, non-destructive agents capable of reliable execution within highly coupled digital ecosystems.

Our major contributions are summarized as follows: 
\begin{itemize}
    \item We introduce DocOps, a rigorously verifiable evaluation framework for end-to-end document manipulation.
    \item We conduct a broad empirical evaluation of leading open- and closed-source models across diverse agentic harnesses, revealing the critical performance bottlenecks of contemporary systems.
    \item Through fine-grained error analysis, we identify pervasive failure modes in document-centric agent systems.
\end{itemize}

\section{Related Work}
\label{sec:related-work}


\paragraph{Autonomous agents.}
Autonomous agents have become central to automating complex digital workflows\citep{ijcai2024p890, NEURIPS2024_5d413e48}. Current research predominantly follows two trajectories. The first focuses on enhancing the intrinsic capabilities of large language models, developing advanced techniques for tool-use~\citep{li-etal-2023-api, qin2023toolllmfacilitatinglargelanguage}, long-context window management~\citep{packer2024memgptllmsoperatingsystems, hu2026memoryageaiagents}, and sophisticated memory mechanisms~\citep{wang-etal-2025-unveiling-privacy, wei2025amemguardproactivedefenseframework}. The second trajectory focuses on harness construction, which provides the scaffolding that encapsulates the execution loop, context manager, and tool registry. Recent work demonstrates that an agent's performance relies heavily on this harness, proving that robust system design is as critical as model intelligence for reliable execution in dynamic environments~\citep{yang2024sweagentagentcomputerinterfacesenable, 10.1145/3712003}.

\paragraph{Document-related benchmarks.}
Existing document benchmarks generally fall into two broad paradigms. The first includes read-only document understanding, such as DocVQA~\citep{Mathew_2021_WACV}, DocBench~\citep{zou2024docbenchbenchmarkevaluatingllmbased}, and OmniDocBench~\citep{Ouyang_2025_CVPR}, as well as format-specific manipulation, such as SheetCopilot~\citep{NEURIPS2023_0ff30c4b}, SpreadsheetBench~\citep{NEURIPS2024_ac840df2}, and PPTArena~\citep{ofengenden2025pptarenabenchmarkagenticpowerpoint}. These benchmarks evaluate information extraction or native editing within relatively focused document and task spaces. The second paradigm evaluates longer office workflows and application orchestration, including OfficeBench~\citep{wang2024officebenchbenchmarkinglanguageagents}, OdysseyBench~\citep{wang2025odysseybenchevaluatingllmagents}, DELEGATE-52~\citep{laban2026llmscorruptdocumentsdelegate}, and FINCH~\citep{dong2026finchbenchmarkingfinance}. OfficeBench and OdysseyBench emphasize application-level workflow completion, DELEGATE-52 measures round-trip degradation under repeated reversible edits, and FINCH evaluates authentic finance workflows with heterogeneous document inputs. Their primary evaluation targets therefore differ from deterministic artifact-level verification that directly checks whether the requested native state is reached while preserving document validity and relevant surrounding state.

To provide a direct and comprehensive comparison, we construct a related-work table~\ref{tab:related_benchmark_comparison} above that contrasts DocOps with prior benchmarks across task scope, evaluation contract, and agent-system analysis, which highlights DocOps’s central novelty: \textit{treating documents as first-class, stateful computational objects, where each operation is evaluated as a state transition that must reach the requested state while preserving native-format validity and relevant out-of-scope document state across heterogeneous formats.} This positions DocOps as a unified framework for evaluating and diagnosing reliable document operations, rather than merely an incremental expansion in formats or task coverage.

\section{DocOps}


DocOps is designed to evaluate whether agents can modify native-format office artifacts while preserving their document-native state. It combines a two-axis task taxonomy(Section~\ref{sec:taxonomy}), controlled task construction(Section~\ref{sec:construction}), and deterministic final-artifact verification(Section~\ref{sec:verifier}).

\subsection{Taxonomy Design}
\label{sec:taxonomy}

Our taxonomy uses two axes to support both fine-grained operation diagnosis and workflow-level reliability analysis. The \textbf{operation axis} targets the distinct editable layers of structured office documents~\citep{10.1117/12.476326}. The \textbf{difficulty axis} measures an agent's ability to maintain state across complex, multi-step operation sequences~\citep{LIU2012553,STANTON200655}

\noindent\textbf{Operation taxonomy.}
The operation axis categorizes atomic document-editing primitives into three families: 1) \emph{Content} evaluates the model's ability to handle textual or numeric semantics, including extracting information, editing text, generating grounded text, computing formulas and reasoning. 2)\emph{Format} focuses on the presentation state of an artifact and tests whether the model can follow formatting instructions, including style consistency, highlighting key words, layout instructions and transferring theme. 3) \emph{Structure} focuses on the organization of document-native objects across formats, such as reordering slides, inserting a cell, modifying outline levels, and operating on native tables or worksheets.

\noindent\textbf{Difficulty taxonomy.}
The difficulty axis measures the depth of planning and state-tracking ability required by a task. \textbf{L1} contains a single atomic edit and tests local execution. \textbf{L2} combines several atomic operations within one artifact and tests short-horizon planning and local dependency modeling. \textbf{L3} requires a document-level workflow over one complex artifact and tests whether the agent can maintain global consistency across multiple steps. \textbf{L4} requires the production or alignment of a final artifact from multiple documents and tests cross-document state tracking and information alignment.

\noindent\textbf{Task package.}
Each task is released as a self-contained Harbor bundle~\citep{merrill2026terminalbenchbenchmarkingagentshard,Harbor_Framework}, containing the source artifact(s), a natural-language instruction, optional document skills, and a deterministic verifier. The verifier directly checks the submitted final artifact, allowing the taxonomy to support both agent execution and reproducible evaluation.

\subsection{Task Construction}
\label{sec:construction}
DocOps is a controlled, preservation-aware synthetic benchmark inspired by practical document-operation requests. To balance practical relevance with reproducible evaluation, we construct DocOps through a four-stage pipeline that produces compact, well-specified tasks with explicit editing scopes and preservation requirements.

\noindent\textbf{Stage 1: Seed collection and filtering.}
We screened 600 candidate sources, including 472 community requests, 125 public workflow references and workflow-level cases, and 3 automation templates. We retained 482 candidates and excluded 118 (19.7\%) that required external services or lacked a concrete offline operation, a clear editing scope, verifiable success conditions, redistribution safety, or sufficient distinction from existing patterns. Community requests primarily inform localized operations, while workflow specifications and automation cases inform longer single-document and cross-document workflows. The retained sources form a candidate pool and do not map one-to-one to the final tasks.

\noindent\textbf{Stage 2: Task formalization.}
We convert each retained seed into a structured specification through standardized prompting and store it in \texttt{task\_metadata.json}, as detailed in Appendix~\ref{app:task-formalization-prompt}. The specification records the taxonomy labels, input and output paths, required operations, editing scope, and preservation requirements. Ambiguous or internally inconsistent specifications are revised before artifact construction.

\noindent\textbf{Stage 3: Source-artifact synthesis.}
Based on each formal specification, we construct a compact native-format source artifact containing the content, structure, and dependencies required by the task. This includes spreadsheet records and formulas, Word sections, slide objects and PDF pages. The resulting artifacts are placed at the prescribed input paths and retain the task-specific state needed to evaluate scoped editing and preservation.

\begin{table*}[t]
\centering
\small
\setlength{\tabcolsep}{4.2pt}
\renewcommand{\arraystretch}{1.15}
\begin{tabular}{@{}>{\raggedright\arraybackslash}m{4.05cm}cccccc@{}}
\toprule
\multirow{2}{*}{\raisebox{-4pt}{\textbf{LLM Agents}}}
& \multirow{2}{*}{\raisebox{-3pt}{\textbf{DocTools}}}
& \multirow{2}{*}{\raisebox{-3pt}{\textbf{Terminus-2}}}
& \multicolumn{2}{c}{\textbf{Codex}}
& \multicolumn{2}{c}{\textbf{Claude Code}} \\
\cmidrule(lr){4-5}
\cmidrule(lr){6-7}
& & &
\textbf{w/ Skill}
& \textbf{w/o Skill}
& \textbf{w/ Skill}
& \textbf{w/o Skill} \\
\midrule
\addlinespace[2pt]
\multicolumn{7}{@{}l}{\textbf{\textit{Closed-source models}}} \\
\addlinespace[1pt]
\hspace{0.6em}GPT-5.5 & \best{0.138} & \best{0.524} & \best{0.671} & \best{0.648} & \na & \na \\

\hspace{0.6em}GPT-5.4 & \second{0.119} & 0.452 & \second{0.662} & \second{0.638} & \na & \na \\

\hspace{0.6em}Claude Sonnet 4.6 & \second{0.119} & 0.419 & \na & \na & \best{0.519} & \best{0.552} \\

\midrule
\addlinespace[2pt]
\multicolumn{7}{@{}l}{\textbf{\textit{Open-source models}}} \\
\addlinespace[1pt]
\hspace{0.6em}DeepSeek-V4-Pro & 0.057 & 0.467 & 0.424 & 0.400 & 0.400 & \second{0.433} \\

\hspace{0.6em}Gemma4-31B & 0.033 & 0.281 & 0.338 & 0.324 & 0.300 & 0.233 \\

\hspace{0.6em}Qwen3.6-27B & 0.095 & \second{0.476} & 0.210 & 0.248 & \second{0.424} & 0.386 \\

\hspace{0.6em}Qwen3.6-35B-A3B & 0.086 & 0.386 & 0.219 & 0.205 & 0.338 & 0.329 \\

\hspace{0.6em}Qwen3.5-122B-A10B & 0.100 & 0.362 & 0.281 & 0.290 & 0.290 & 0.295 \\

\hspace{0.6em}Qwen3.5-27B & 0.114 & 0.362 & 0.295 & 0.271 & 0.371 & 0.300 \\

\hspace{0.6em}Qwen3.5-35B-A3B & 0.105 & 0.300 & 0.176 & 0.157 & 0.295 & 0.262 \\

\hspace{0.6em}Qwen3.5-9B & 0.081 & 0.233 & 0.086 & 0.062 & 0.176 & 0.181 \\

\hspace{0.6em}GLM-4.5-Air & 0.071 & 0.162 & \na & \na & 0.167 & 0.129 \\

\bottomrule
\end{tabular}
\caption{Overall task pass rates across models and execution harnesses. Bold and underlined entries mark the best and second-best results in each column, respectively. ``\na{}'' denotes unavailable combinations due to protocol incompatibility, not evaluated failures.}
\label{tab:overall_results}
\end{table*}


\noindent\textbf{Stage 4: Iterative human review.}
One main human reviewer, a PhD-level researcher specializing in document operations, iteratively compares each instruction with its source artifacts and metadata. The review checks instruction clarity, edit scope completeness, taxonomy alignment, native-file integrity, and input and output path consistency. Tasks that fail these checks are returned to the relevant earlier stage for revision or discarded. For example, an early PowerPoint candidate instructed the agent to modify Slide 2, while the source presentation contained only one slide. Human review detected this mismatch between instructions and artifacts, and the candidate was rejected and corrected before inclusion.

Following this pipeline, DocOps contains \textbf{210} tasks, comprising \textbf{50} L1 atomic-edit tasks, \textbf{40} L2 compositional tasks, \textbf{60} L3 single-document workflow tasks, and \textbf{60} L4 cross-document workflow tasks. Each task is packaged as a self-contained Harbor package that contains source artifact(s), a natural language instruction, optional document skills, and a deterministic verifier. Detailed benchmark statistics are provided in Appendix~\ref{app:task_statistics}.

\subsection{Verifier Construction and Fidelity}
\label{sec:verifier}

Each task employs a deterministic in-container verifier that inspects the final artifact through document-native libraries. The resulting scoring procedure is deterministic, offline, and independent of LLM-as-a-judge evaluation. To accommodate multiple valid execution paths, the verifier avoids whole-artifact exact matching and instead uses three targeted predicate types. 1) \textbf{Structural predicates} validate underlying native states, such as executable formulas and outline hierarchies, to detect structural corruption that may be invisible in rendered outputs. 2) \textbf{Linguistic anchors} verify required textual content using task-specific diagnostic keywords, maintaining content accuracy while allowing valid linguistic variation. 3) \textbf{Preservation predicates} check whether specified out-of-scope elements, such as protected styles and unmodified worksheets, remain intact.

\noindent\textbf{Verifier fidelity.}
We directly assess verifier fidelity through a human audit and a mutation-based stress test. For the human audit, a computer science PhD candidate experienced in document operations inspected the task instruction, source artifacts, and submitted artifact for 128 sampled verifier decisions and assigned an artifact-level pass or fail judgment. The verifier agrees with these judgments in 122 cases (\textbf{95.31\%}), with three false passes and three false fails. Across 180 controlled mutations of 36 verified outputs, it detects \textbf{174 violations (96.67\%)} spanning requested content, native structure, preservation, missing outputs, and file corruption. Together, these results support verifier fidelity on both sampled outputs and controlled corruptions. Appendix~\ref{app:verifier_fidelity} provides the full protocol, category-level results, and representative disagreement cases.


\section{Experimental}
\label{sec:experiments-setup}

\subsection{Setup}
\subsubsection{Models}

We evaluate both closed- and open-source model groups, as organized in Table~\ref{tab:overall_results}. Specifically, we include GPT-5.5~\citep{openai2026gpt55}, GPT-5.4~\citep{openai2026gpt54}, and Claude Sonnet 4.6~\citep{anthropic2026sonnet46}. The open-source group includes DeepSeek-V4-Pro~\citep{deepseekai2026deepseekv4}, Qwen3.6-27B~\citep{qwen36_27b}, Qwen3.6-35B-A3B~\citep{qwen36_35b_a3b}, Qwen3.5-122B-A10B, Qwen3.5-27B, Qwen3.5-35B-A3B, Qwen3.5-9B~\citep{qwen35blog}, Gemma4-31B~\citep{gemmateam2026gemma4}, and GLM-4.5-Air~\citep{5team2025glm45agenticreasoningcoding}. All evaluated models were publicly accessible at the time of evaluation through provider APIs or publicly available checkpoints. Their exact model IDs, access routes, and serving configurations are provided in Appendix~\ref{sec:app_model_serving}.

\vspace{-4pt}
\subsubsection{Harnesses}
\label{sec:conditions}
We evaluate four execution harnesses that instantiate different interfaces through which the same model can operate on document tasks. These harnesses cover the main interface regimes from constrained tool use to unrestricted programming. 1) \textbf{DocTools:} custom structured document APIs without shell access. 2) \textbf{Terminus-2:} a stateful Bash-based harness with file-system access~\citep{harbor2026terminus2}. 3) \textbf{Claude Code:} an interactive coding harness with file-system feedback~\citep{anthropic2025claudecode}. 4) \textbf{Codex:} a CLI-based harness for open script execution~\citep{openai2025codexcli}.

\subsubsection{Skill Setting}

For Claude Code and Codex, we evaluate both \emph{w/ skill} and \emph{w/o skill} settings. The skill setting uses the official Anthropic document skills for \texttt{xlsx}, \texttt{docx}, \texttt{pptx}, and \texttt{pdf}, released in the Anthropic skills repository~\citep{anthropic_skills_github}. In the w/o skill setting, these document-skill modules are removed or made unavailable while keeping the same underlying harness.

\subsubsection{Evaluation Protocol}

All experiments are executed through Harbor~\citep{merrill2026terminalbenchbenchmarkingagentshard,Harbor_Framework}, which manages task containers, agent execution and verifier invocation. Each setting is evaluated on the full 210-task suite. A task is counted as successful only if the agent writes the required output artifact to the designated submission path and the deterministic verifier returns success. Runtime errors, missing outputs, verifier failures and timeouts are counted as failures.

\subsubsection{Metrics}

We report pass rate as the primary metric:
\vspace{-4pt}
\[
\mathrm{PassRate}(m,h)
=
\frac{1}{|\mathcal{T}|}
\sum_{t\in\mathcal{T}}
\{V(m,h,t)=1\},
\]
where \(V(m,h,t)\) is the verifier result for model \(m\), harness \(h\), and task \(t\), and \(\mathcal{T}\) is the 210-task suite. We also report the same metric over subsets grouped by difficulty level, document format, and operation family.

\subsection{Results}


\subsubsection{Main Results}

\textbf{Frontier agents exhibit significant limitations in document manipulation tasks.}
Table~\ref{tab:overall_results} reports the overall pass rates in all the model and harness configurations evaluated. As shown in Table~\ref{tab:overall_results}, the highest-performing setup (GPT-5.5 with Codex and skills) achieves only a pass rate of 0.671, still failing nearly one-third of the tasks. Furthermore, model capability alone does not dictate success, harness selection is critical. Claude Sonnet 4.6 achieves a competitive pass rate of 0.552 under Claude Code without skills. Terminus-2 also substantially improves several open-source models over direct document-tool use, with Qwen3.6-27B reaching 0.476 under Terminus-2. These results show that document-operation performance depends on the combination of different model capability and execution framework, while even the best current configuration remains far from fully reliable.

\begin{figure}[t]
\centering
\setlength{\belowcaptionskip}{-0.4cm}
\includegraphics[width=0.9\columnwidth]{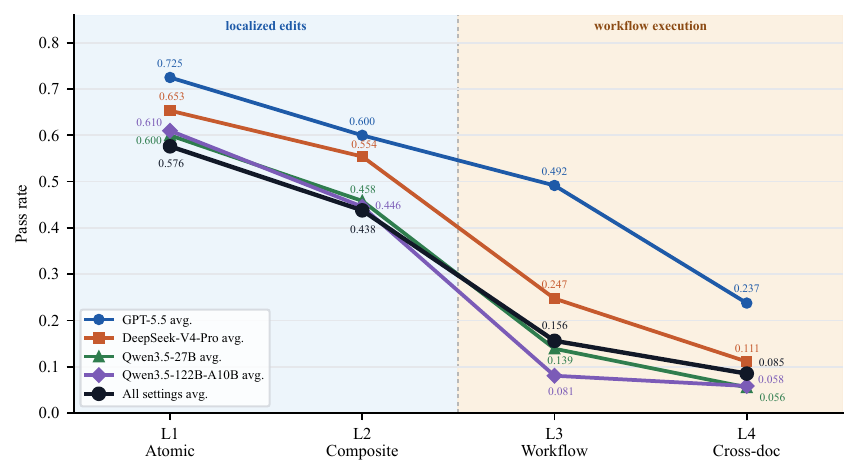}
\caption{Pass rates across difficulty levels for representative models averaged over their available harness and skill settings. Shaded regions distinguish localized edits from workflow-level tasks, and the black curve reports the average over all evaluated settings. 
}
\label{fig:difficulty_gradient}
\end{figure}

\noindent\textbf{Performance collapses once tasks reach workflow-level difficulty.} Figure~\ref{fig:difficulty_gradient} shows that most agents perform relatively well on localized and compositional L1/L2 tasks, but degrade sharply once tasks require workflow-level state tracking and cross-step dependency management.For example, GPT-5.5 decreases from 0.725 on L1 to 0.237 on L4 when averaged over its available harness settings, with similar trends observed for DeepSeek-V4-Pro and the open-source models. Overall, this shared degradation suggests that the central challenge lies not in localized document operations, but in maintaining document state and cross-step dependencies throughout long-horizon workflows.

\begin{table}[t]
\centering
\scriptsize
\setlength{\belowcaptionskip}{-0.4cm}
\setlength{\tabcolsep}{3.2pt}
\renewcommand{\arraystretch}{1.08}
\begin{tabular}{lrrrr}
\toprule
Harness & s/task & min/run & tok/task & tok/run \\
\midrule
DocTools & 74.5 & 260.8 & 31.5K$^\dagger$ & 6.62M$^\dagger$ \\
Terminus-2 & 160.7 & 562.3 & 141.3K & 29.68M \\
Codex w/ Skill & 109.9 & 384.7 & 169.3K$^\dagger$ & 35.56M$^\dagger$ \\
Codex w/o Skill & 94.5 & 330.6 & 152.8K$^\dagger$ & 32.10M$^\dagger$ \\
Claude Code w/ Skill & 182.6 & 639.0 & 664.9K & 139.64M \\
Claude Code w/o Skill & 164.4 & 575.2 & 590.0K & 123.89M \\
\bottomrule
\end{tabular}
\caption{Average harness cost statistics. $^\dagger$ denotes calibrated token estimates.}
\label{tab:harness_cost}
\end{table}

\begin{figure*}[t]
\centering
\setlength{\belowcaptionskip}{-0.6cm}
\includegraphics[width=0.85\textwidth]{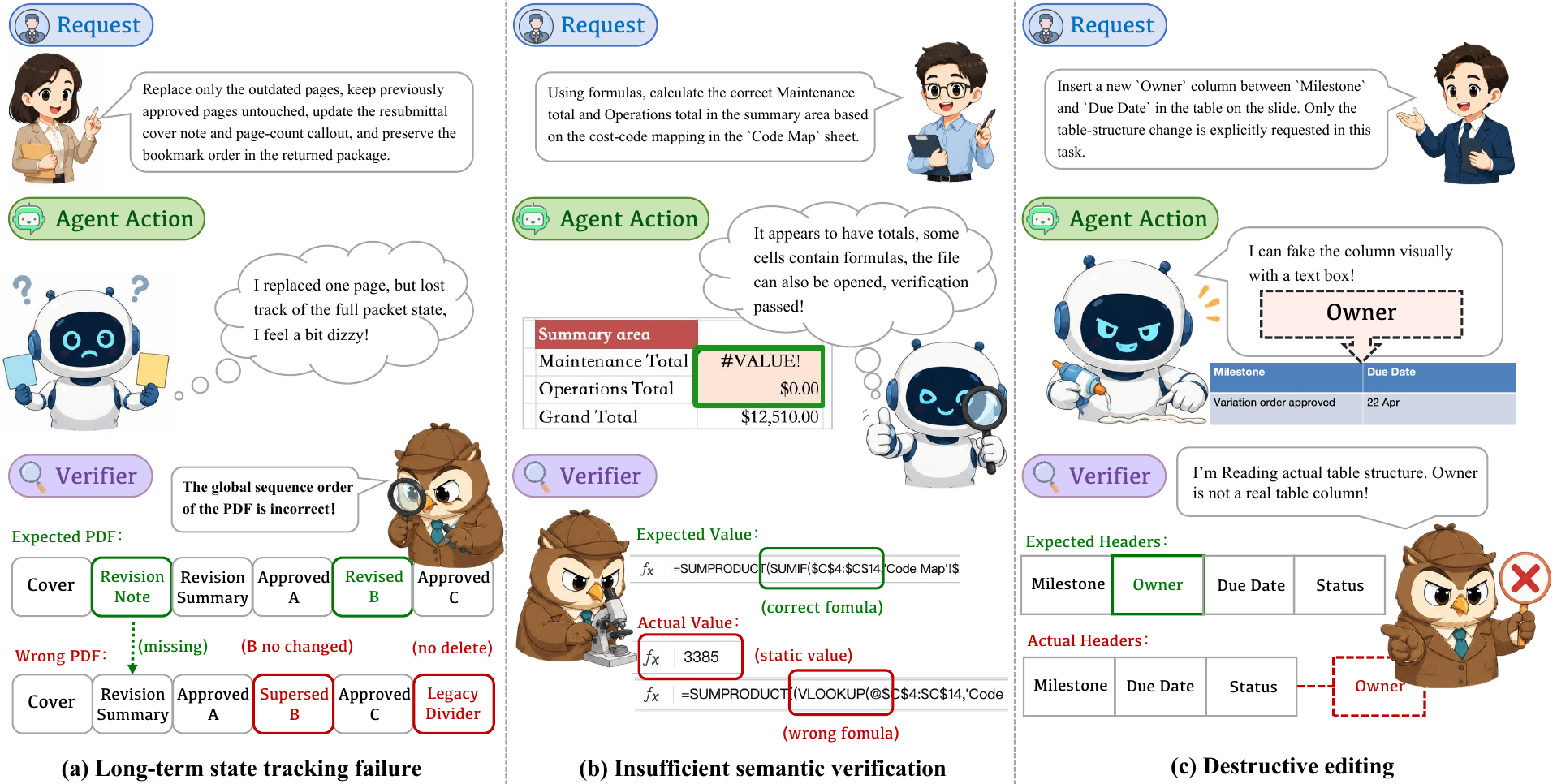}
\caption{
\textbf{Representative trajectory-level failure cases} showing typical agent behaviors under three failure modes:
(a) long-term state tracking failure. (b) semantic verification gap. (c) destructive editing. }
\label{fig:failure_cases}
\end{figure*}

\noindent\textbf{Richer execution scaffolding tends to carry greater resource overhead, with Claude Code recording the highest cost in both runtime and token usage.}
To characterize execution efficiency beyond task success, Table~\ref{tab:harness_cost} reports average runtime and token usage for each harness. DocTools incurs the lowest cost in both dimensions, consistent with its constrained document-tool interface. Compared with Terminus-2, Codex completes tasks substantially faster but consumes moderately more tokens, suggesting that its execution loop exchanges additional model interaction for lower latency. Claude Code exhibits a different profile, recording the highest runtime and token usage across all settings. Enabling skills further increases both costs under Codex and Claude Code. Overall, the execution interface directly shapes the resource profile of a harness, while skill injection introduces additional overhead whose cost must be considered alongside its non-uniform performance effects.

\subsubsection{Detailed Analysis}

\noindent\textbf{Workflow-level difficulty depends not only on the number of operations, but also on how tightly their underlying document states are coupled.}
Table~\ref{tab:format_level_cv} reports pass rates by document format and difficulty level with 95\% bootstrap confidence intervals. We further compute FormatCV to quantify cross-format performance variation:
\vspace{-6pt}
\[
\mathrm{FormatCV}(\ell)=
\frac{\mathrm{Std}_{d\in\mathcal{D}}(\mathrm{Acc}_{\ell,d})}
{\mathrm{Mean}_{d\in\mathcal{D}}(\mathrm{Acc}_{\ell,d})},
\]
where $\mathcal{D}=\{\textsc{Excel}, \textsc{Word}, \textsc{PPT}, \textsc{PDF}\}$.
The format-level breakdown shows that workflow degradation is highly uneven. Excel shifts from the highest pass rate at L1 to the lowest at L3 and exhibits the largest decline. This sharp degradation highlights the challenge of jointly maintaining formulas, cell references, sheet order, and hidden states, where a local error can propagate through the workbook. In contrast, PDF retains the highest L3 pass rate, showing that agents remain more reliable when workflow operations involve comparatively modular page-level state changes. Accordingly, FormatCV increases from 0.120 at L1 to 0.436 at L3, with non-overlapping bootstrap confidence intervals. These results suggest that the coupling of native document states, rather than operation count alone, is a key source of workflow-level difficulty.

\begin{table}[t]
\centering
\scriptsize
\setlength{\tabcolsep}{1.7pt}
\setlength{\belowcaptionskip}{-0.4cm}
\renewcommand{\arraystretch}{2}

\newcommand{\rowlabel}[1]{%
\begin{tabular}[c]{@{}l@{}}#1\end{tabular}%
}

\newcommand{\cellci}[2]{%
\begingroup
\renewcommand{\arraystretch}{0.92}%
\begin{tabular}[c]{@{}c@{}}
$#1$\\[-1.0pt]
{\tiny$[#2]$}
\end{tabular}%
\endgroup
}

\newcommand{\chgbase}[2]{%
\begingroup
\renewcommand{\arraystretch}{0.90}%
\begin{tabular}[c]{@{}c@{}}
{\tiny\textcolor{#1}{$#2$}}\\[-1.1pt]
{\scriptsize$\longrightarrow$}
\end{tabular}%
\endgroup
}

\newcommand{\decchg}[1]{\chgbase{red!70!black}{-#1}}
\newcommand{\incchg}[1]{\chgbase{green!45!black}{+#1}}

\begin{tabular}{@{}lccccc@{}}
\toprule
\noalign{\vskip -2pt}
\textbf{Format}
& \textbf{L1}
&
& \textbf{L2}
&
& \textbf{L3} \\[-2pt]
\midrule
\rowlabel{Excel}
& \cellci{\mathbf{0.671}}{0.566,0.773}
& \decchg{0.238}
& \cellci{0.433}{0.247,0.623}
& \decchg{0.374}
& \cellci{0.058}{0.031,0.085} \\

\rowlabel{Word}
& \cellci{0.551}{0.382,0.699}
& \decchg{0.018}
& \cellci{\mathbf{0.533}}{0.391,0.656}
& \decchg{0.330}
& \cellci{0.203}{0.140,0.275} \\

\rowlabel{PPT}
& \cellci{0.587}{0.438,0.715}
& \decchg{0.187}
& \cellci{0.400}{0.247,0.545}
& \decchg{0.271}
& \cellci{0.129}{0.070,0.201} \\

\rowlabel{PDF}
& \cellci{0.480}{0.325,0.641}
& \decchg{0.095}
& \cellci{0.386}{0.256,0.505}
& \decchg{0.153}
& \cellci{\mathbf{0.233}}{0.156,0.316} \\

\midrule
\noalign{\vskip -2pt}
\rowlabel{FormatCV}
& \cellci{0.120}{0.056,0.279}
& \incchg{0.011}
& \cellci{0.131}{0.061,0.342}
& \incchg{0.304}
& \cellci{0.436}{0.321,0.619} \\[-2pt]
\bottomrule
\end{tabular}

\caption{Average pass rates by document type and difficulty level with 95\% bootstrap confidence intervals. Inter-column arrows show changes from the previous difficulty level; FormatCV is computed across types.}
\label{tab:format_level_cv}
\end{table}

\begin{figure}[t]
\centering
\setlength{\belowcaptionskip}{-0.5cm}
\includegraphics[width=1\columnwidth]{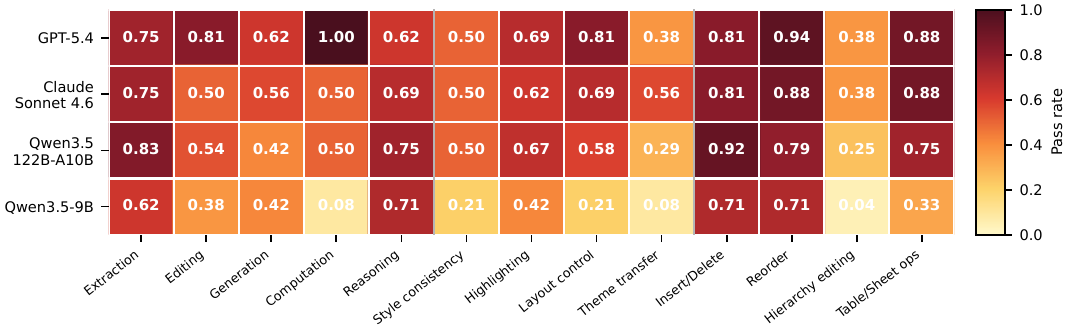}
\caption{Atomic operation accuracy for representative models on L1 tasks. Scores are averaged over the available harness settings for each model, the full-model heatmap is reported in Appendix~\ref{app:full_operation_heatmap}.}
\label{fig:model_operation_heatmap_rep}
\end{figure}

\noindent\textbf{Current agents consistently struggle with atomic operations that require document-native state control.}
Figure~\ref{fig:model_operation_heatmap_rep} reports L1 atomic-operation accuracy for representative models. The heatmap shows that operations that rely on localized content access or object movement, such as extraction, reordering, and insert/delete, are relatively reliable. In contrast, hierarchy editing and theme transfer remain difficult, and computation and style consistency exhibit larger variation across model strengths. This suggests that the bottleneck operations for document agents are those that require preserving or updating document-native state, including formulas, styles, themes, and structural hierarchies, rather than simply reading content or moving visible objects.

\subsubsection{Key Failure Modes}

Despite aggregate pass rates, we analyze failures using triggered verifier assertions and execution trajectories. The results show that agents often produce locally plausible edits, yet struggle to maintain document state, verify task-specific semantics, and preserve native document structure. We define three key failure modes as follows.

\noindent\textbf{Long-term state tracking failure: Agents lose global document state during multi-step editing.}
We define this failure mode as cases where the violated assertions concern persistent document state, such as page, slide, or sheet order, preserved objects, hidden sheets, or cross-document mappings. As shown in Figure~\ref{fig:failure_cases}(a), the agent performs local PDF page edits, but the final document still contains stale or misplaced components. This suggests that the agent fails to maintain a consistent global document plan after each edit, and therefore cannot account for the full document state.

\noindent\textbf{Semantic verification gap: Agents accept visually plausible outputs without verifying executable semantics.}
We define this failure as cases where an output file is produced but violates task-specific semantic constraints, such as required formulas, value ranges, hierarchy, or structural conditions. In Figure~\ref{fig:failure_cases}(b), the spreadsheet contains plausible static values or seemingly complete formulas, but the key cells contain missing or incorrect formulas. This indicates a clear gap between generic read-back verification and the precise checks required by document-centric workflows.

\noindent\textbf{Destructive editing: Agents complete local edits by damaging native document structure.}
We define this failure mode as cases where the output breaks document-native invariants, such as formulas, validation rules, styles, heading hierarchy, native tables, bookmarks, or protected reference content. In Figure~\ref{fig:failure_cases}(c), the PowerPoint table appears visually patched, but the underlying native table structure is not correctly updated. Such failures show that agents often treat documents as surface layouts rather than structured editable artifacts.

\begin{figure}[t]
\centering
\setlength{\belowcaptionskip}{-0.4cm}
\includegraphics[width=0.9\columnwidth]{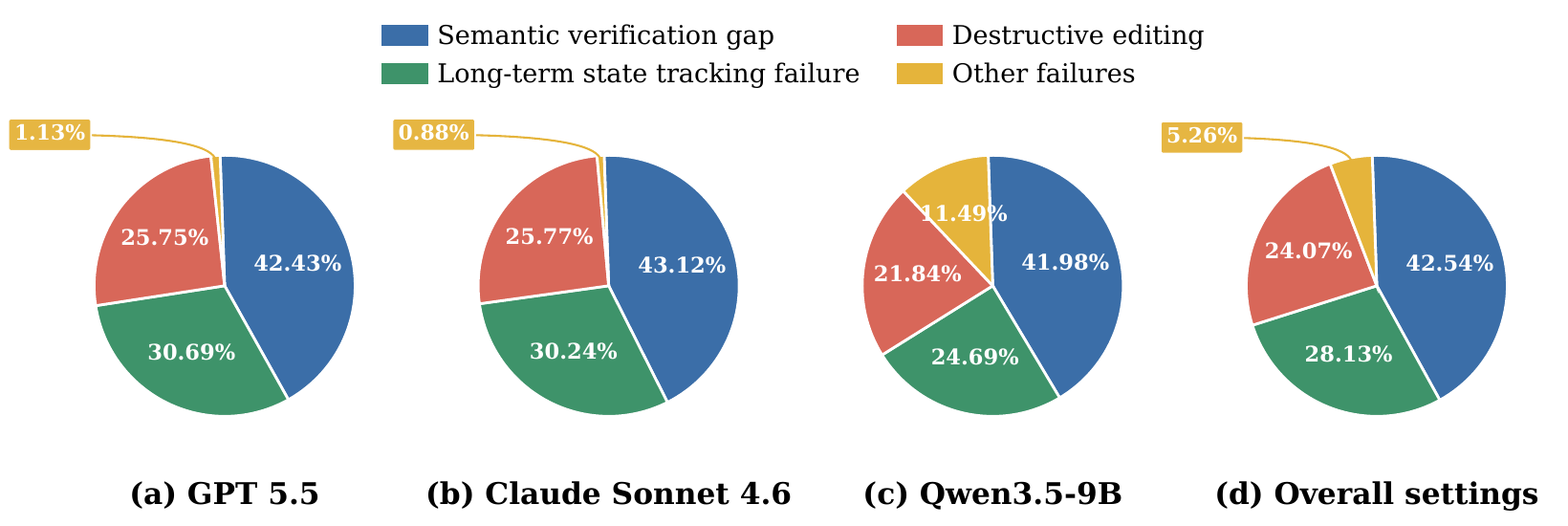}
\caption{
Distribution of different failure modes across representative settings and the overall aggregate.}
\label{fig:failure_modes_representative}
\end{figure}

To quantify how these issues affect task failures, we extract verifier-grounded signals from failed runs and group them by failure mode. Figure~\ref{fig:failure_modes_representative} shows that semantic verification gaps dominate across representative models and overall settings, while state tracking failures and destructive editing remain substantial. These patterns suggest that failures stem not only from instruction-following errors but also from weaknesses in semantic checking, state maintenance, and non-destructive editing.


\subsubsection{Impact of Harness and Skills}

DocOps shows that the execution harness is a major part of the document agent. The same underlying model can behave very differently depending on whether the harness exposes an open programming environment, a constrained tool interface, or additional procedural skills.

\begin{figure}[t]
\centering
\setlength{\abovecaptionskip}{-0.1cm}
\setlength{\belowcaptionskip}{-0.4cm}
\includegraphics[width=0.9\linewidth]{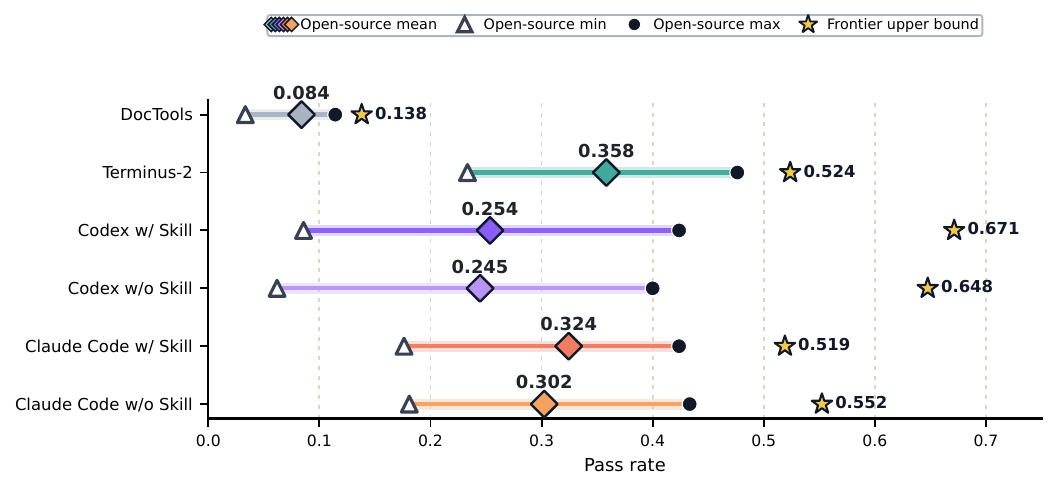}
\caption{Harness-level pass-rate ranges over open-source models with complete harness coverage, excluding GLM-4.5-Air due to missing Codex results. Triangles, circles, and diamonds denote minima, maxima, and means, while stars mark the best evaluated closed-source configuration for each harness.}
\label{fig:section4_harness_range}
\end{figure}

\noindent\textbf{Harness effectiveness depends strongly on the paired model, and open-source configurations remain below their frontier counterparts.}
Figure~\ref{fig:section4_harness_range} reports the mean and range of harness-level pass rates across open-source models. The broad ranges show that the same execution framework can yield substantially different outcomes depending on the model. Terminus-2 achieves the strongest open-source aggregate with comparatively stable performance, whereas Codex and Claude Code exhibit greater variation. Thus, greater framework flexibility does not produce uniform gains, since its effectiveness depends on the model's ability to use the execution environment reliably.

The frontier reference markers further illustrate this interaction. Under Codex and Claude Code, the best observed closed-source configurations substantially exceed the corresponding open-source ranges, showing that these harnesses attain much stronger performance when paired with frontier or even official models. Overall, document-operation performance is shaped by the interaction between model capability and execution framework, while current open-source configurations remain considerably behind the strongest evaluated frontier configurations.

\begin{figure}[t]
\centering
\setlength{\belowcaptionskip}{-0.4cm}
\includegraphics[width=0.88\linewidth]{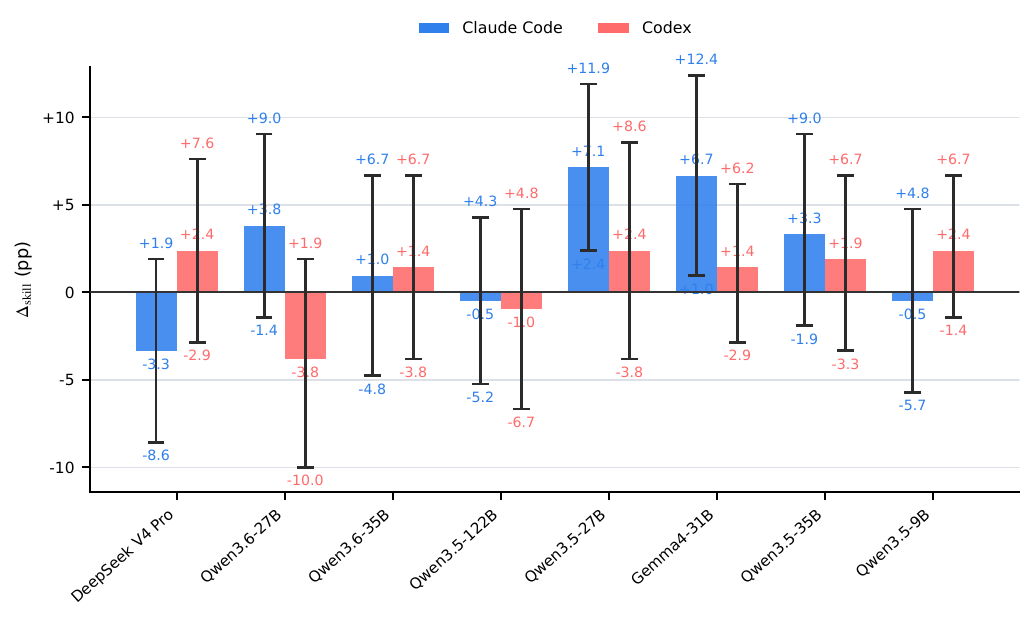}
\caption{Overall pass-rate change from skill injection on open-source models. Bars report $\Delta_{\mathrm{skill}}=\mathrm{Acc}_{\mathrm{w/}}-\mathrm{Acc}
_{\mathrm{w/o}}$
under Claude Code and Codex. Positive values indicate gains and negative values indicate regressions.}
\label{fig:section4_skill_delta_overall}
\end{figure}

\noindent\textbf{Skill injection does not consistently improve document-operation performance and can even constrain effective problem solving.}
Figure~\ref{fig:section4_skill_delta_overall} reports paired pass-rate differences between skill-enabled and no-skill settings for selected open-source models under Claude Code and Codex. For each model and harness pair, we perform 10,000 paired bootstrap resamples over the common task set and report 95\% confidence intervals. Under Claude Code, Qwen3.5-27B and Gemma4-31B improve by 7.1 and 6.7 percentage points, respectively, with intervals entirely above zero. However, DeepSeek-V4-Pro and Qwen3.5-9B show negative point estimates, and several intervals overlap zero. Under Codex, the effects are generally smaller and less certain: Qwen3.6-27B decreases by 3.8 points, while most other intervals cross zero. These results show that static skill injection does not consistently improve document-operation performance and should be treated as an empirical design choice. Fine-grained trajectory analysis further suggests that overly rigid adherence to prescribed procedures can limit adaptation and alternative problem-solving strategies across diverse tasks. Appendix~\ref{app:skill-delta-full} provides the full breakdown by difficulty level and operation family.

\section*{Conclusion}
In conclusion, DocOps establishes a controlled, preservation-aware benchmark for evaluating autonomous agents on native-format artifacts. Across the evaluated model–harness configurations, performance declines substantially as tasks require longer-horizon state tracking, cross-step dependency management, and preservation of document-native structure. Our analysis identifies recurring failures in maintaining global state, verifying task-specific semantics, and preserving native structure. Ultimately, DocOps provides a reproducible diagnostic framework and empirical foundation for guiding the development of next-generation agents capable of maintaining global consistency and avoiding destructive modifications across complex digital artifacts.

\section*{Limitations}
1) DocOps focuses on deterministic, offline document-editing tasks and does not cover workflows requiring live external services, collaborative editing, or interactive user clarification. 
2) Because complex document tasks require structurally valid artifacts, clear editing scopes, and manual review of both instructions and generated files, scaling the benchmark is substantially more labor-intensive than collecting read-only document examples. DocOps currently contains 210 tasks. In future work, we will continue expanding the suite by adding more document domains, larger workflow packets, and additional format-specific operations while preserving deterministic verification and human quality control.
3) Token-cost comparisons across harnesses should be interpreted with care because different agent runtimes expose usage statistics with different fidelity.


\bibliography{custom}

\appendix

\section*{Appendix}
\label{sec:appendix}

\section{Task Statistics and Full Task List}
\label{app:task_statistics}
\begin{figure}[t]
\centering
\includegraphics[width=\columnwidth]{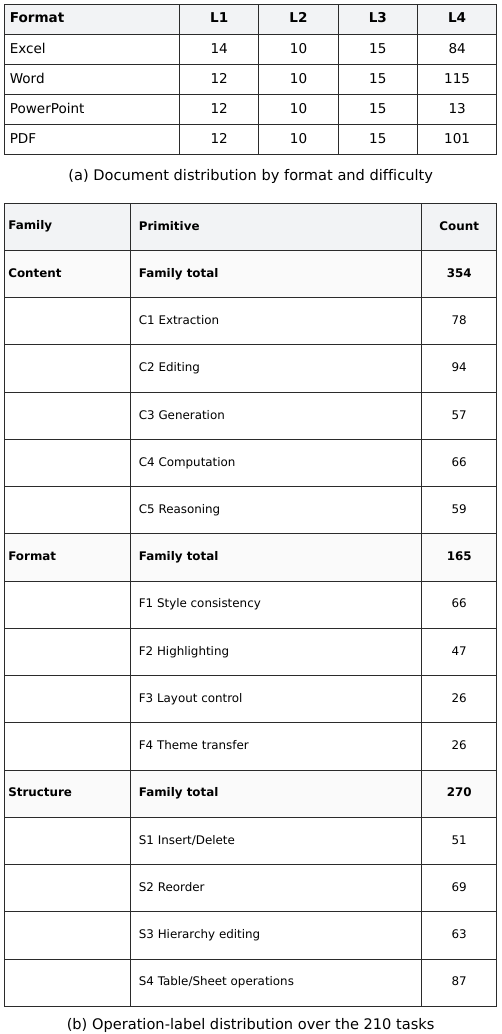}
\caption{DocOps benchmark statistics.}
\label{fig:app_benchmark_stats}
\end{figure}

Figure~\ref{fig:app_benchmark_stats} presents two complementary views of the expanded 210-task DocOps benchmark. Panel~(a) reports the number of source artifacts by format and difficulty level, rather than the number of tasks. This distinction is particularly important for L4 cross-document tasks, since a single task may contain multiple source artifacts in different formats. Panel~(b) shows the distribution of operation labels across the full benchmark under the Content, Format, and Structure taxonomy. Because composite and workflow-level tasks may involve multiple primitives, these values represent label occurrences rather than mutually exclusive task counts. Accordingly, each family total is the sum of its primitive-label occurrences.

Table~\ref{tab:app_full_task_list} provides the complete listing of all 210 tasks and is placed at the end of the appendix for readability. For each task, we report the task identifier, difficulty level, operation labels, normalized input formats, and required output format. For L4 cross-document tasks, the input-format column includes all unique source formats contained in the task environment.

\begin{figure*}[t]
\centering
\fbox{
\begin{minipage}{0.96\textwidth}
\scriptsize
\textbf{Task Formalization Prompt.}
Given an informal seed describing a document-editing need, convert it into a
structured
\texttt{task\_metadata.json} specification for a self-contained benchmark
task.

\textbf{Input:} \texttt{\{SEED\_TEXT\}}

\textbf{Output:} Return only valid JSON using the fields below.

\textbf{Common fields for all tasks:}
\texttt{task\_id};
\texttt{task\_dir\_name};
\texttt{doc\_type};
\texttt{primary\_modality};
\texttt{difficulty};
\texttt{verifier\_mode};
\texttt{user\_task\_description};
\texttt{single\_step\_benchmark\_instruction};
\texttt{atomic\_scope};
\texttt{expected\_output\_type};
\texttt{notes};
\texttt{input\_path};
\texttt{output\_path};
\texttt{output\_filename};
\texttt{output\_extension};
\texttt{output\_kind}.

\textbf{Fields for L1 atomic tasks:}
\texttt{atomic\_operation};
\texttt{source\_doc\_filename};
\texttt{source\_doc\_path};
\texttt{task\_description\_filename};
\texttt{task\_description\_path};
\texttt{inspired\_by}.

\textbf{Fields for composite and workflow tasks:}
\texttt{atomic\_operations};
\texttt{num\_atomic\_operations};
\texttt{composition\_type};
\texttt{composition\_pattern};
\texttt{source\_type}.

\textbf{Additional field for cross-document tasks:}
\texttt{supporting\_input\_paths}.

\textbf{Optional field when the final state is fixed by the task design:}
\texttt{verifier\_expectations}.

\textbf{Constraints:}
The task must require editing a real document artifact; be solvable offline
inside a
self-contained container; have a clear required output path; preserve out-of-
scope content
specified in \texttt{atomic\_scope}; and avoid subjective requests such as
purely aesthetic
improvement. Use \texttt{notes} to record latent issues or task-specific
preservation
requirements that should be reflected in the source artifact.
\end{minipage}
}
\caption{Prompt template used to formalize informal document-editing seeds into the metadata schema used by DocOps.}
\label{fig:task-formalization-prompt}
\end{figure*}

\section{Task Formalization Prompt}
\label{app:task-formalization-prompt}

Figure~\ref{fig:task-formalization-prompt} shows the prompt template used in Stage 2 of the construction pipeline. The goal of this prompt is not to generate the final document artifact directly, but to convert an informal document-editing seed into a structured metadata specification. This metadata
acts as the task-level contract for later artifact synthesis and review: it fixes the document format, difficulty level, operation labels, input and output paths, user-facing instruction, editing scope, expected output type, and task-specific notes. For composite, workflow, and cross-document tasks, the same schema additionally records the operation sequence and supporting input files. 

\section{Harness Cost Accounting}
\label{app:harness-cost-accounting}

\begin{figure*}[t]
\setlength{\belowcaptionskip}{-0.4cm}
\centering
\includegraphics[width=\textwidth]{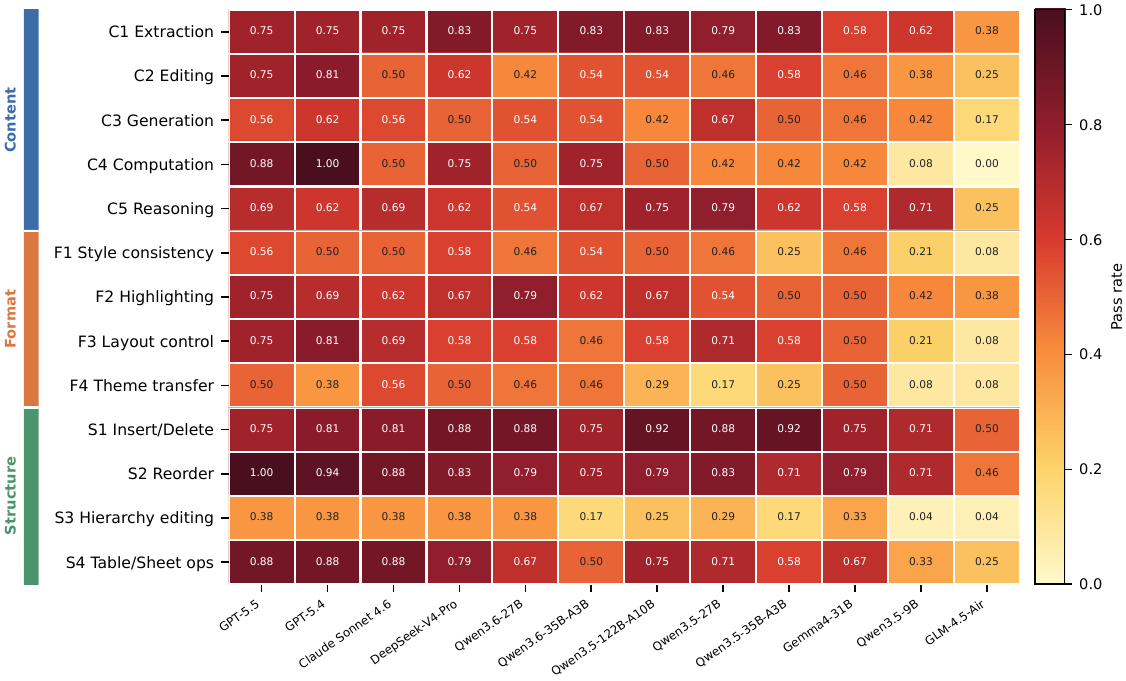}
\caption{Atomic operation accuracy by model on L1 tasks for all evaluated models. Scores are averaged over the available harness settings for each model.}
\label{tab:full_operation_heatmap}
\end{figure*}

Table~\ref{tab:harness_cost} reports average runtime and token usage at the harness level. For all harnesses, runtime is taken directly from Harbor timing records, which measure each task from start to termination and are independent of harness-specific token accounting. This timing includes agent interaction, tool execution, file operations, verifier invocation, and timeout duration when applicable. We report \textit{s/task} as the mean seconds per task across valid settings for each harness, and \textit{min/run} as the corresponding full-run time.

Token usage counts prompt/input and completion/output tokens across the full interaction trajectory. Unlike runtime, token accounting may differ across harnesses because different runtimes expose different usage metadata. Specifically:

\noindent$\bullet$ \textbf{DocTools.} We use the agent-reported usage fields when available. For trajectories where the LLM-call records preserve the visible interaction but do not expose reliable token-usage metadata, we estimate token usage from the visible LLM-call trajectory and calibrate it against DocTools runs with reliable usage accounting.

\noindent$\bullet$ \textbf{Terminus-2.} We compute token usage directly from the recorded usage fields over the full interaction trajectory, without additional calibration.

\noindent$\bullet$ \textbf{Codex.} Some Codex trajectories preserve the full visible interaction history but do not expose reliable server-side usage fields through the runtime interface. For these cases, we estimate token usage from the visible Codex trajectory and calibrate it against Codex runs with reliable token accounting. Values marked with $^\dagger$ in Table~\ref{tab:harness_cost} denote these calibrated token estimates.

\noindent$\bullet$ \textbf{Claude Code.} We compute token usage directly from the recorded usage fields over the full interaction trajectory, without additional calibration.

To keep harness averages comparable, each row is computed only over model settings for which the corresponding harness produced valid run records under the reported experimental configuration. Settings without valid results are excluded because including them would mix unavailable or invalid executions with actual harness cost, thereby distorting the runtime and token averages.

\section{Full operation-level capability map.}
\label{app:full_operation_heatmap}
Figure~\ref{tab:full_operation_heatmap} reports the complete operation-level heatmap for all evaluated models. Each cell gives the average pass rate on L1 atomic tasks for one model and one atomic operation, averaged over the available harness settings for that model. This full view complements the representative heatmap in the main text and confirms that the same bottleneck operations recur across the broader model set. In particular, operations requiring document-native state control, such as hierarchy editing, theme transfer, computation, and style consistency, remain substantially less reliable than localized content access or object movement operations.

\section{Full Skill-Injection Results}
\label{app:skill-delta-full}
Table~\ref{tab:section4_skill_delta_full} reports the full skill-injection deltas for Claude Code and Codex. We
report the overall deltas by difficulty level (L1 to L4) and the deltas by operation family: Content (C), Format (F) and Structure (S). Positive values indicate that enabling skills improves pass rate on the corresponding subset, while negative values indicate regressions. Zero values indicate that the
with-skill and no-skill settings achieve the same pass rate on that subset. Within each harness block and column, boldface marks the largest positive and largest negative delta, highlighting the strongest gains and regressions caused by skill injection.

\begin{table*}[t]
\centering
\small
\setlength{\tabcolsep}{3.2pt}
\renewcommand{\arraystretch}{1.08}

\resizebox{\linewidth}{!}{
\begin{tabular}{lrrrrrrrr}
\toprule
Model & Overall & L1 & L2 & L3 & L4 & C & F & S \\
\midrule
\rowcolor{gray!12}
\multicolumn{9}{c}{\textbf{Claude Code}} \\
Claude Sonnet 4.6 & \textbf{\loss{-0.033}} & \gain{+0.040} & \textbf{\loss{-0.175}} & \textbf{\loss{-0.017}} & \loss{-0.017} & \textbf{\loss{-0.036}} & \textbf{\loss{-0.045}} & \textbf{\loss{-0.044}} \\
DeepSeek-V4-Pro & \loss{-0.033} & \textbf{\loss{-0.060}} & \loss{-0.100} & \gain{+0.050} & \textbf{\loss{-0.050}} & \loss{-0.012} & \loss{-0.015} & \loss{-0.032} \\
Gemma4-31B & \gain{+0.067} & \gain{+0.100} & \gain{+0.075} & \textbf{\gain{+0.067}} & \gain{+0.033} & \textbf{\gain{+0.073}} & \loss{-0.008} & \textbf{\gain{+0.076}} \\
Qwen3.6-27B & \gain{+0.038} & \neutral{+0.000} & \gain{+0.050} & \gain{+0.033} & \textbf{\gain{+0.067}} & \gain{+0.055} & \gain{+0.038} & \gain{+0.051} \\
Qwen3.6-35B-A3B & \gain{+0.010} & \gain{+0.020} & \gain{+0.025} & \gain{+0.017} & \loss{-0.017} & \neutral{+0.000} & \neutral{+0.000} & \gain{+0.006} \\
Qwen3.5-122B-A10B & \loss{-0.005} & \loss{-0.020} & \loss{-0.025} & \gain{+0.033} & \loss{-0.017} & \gain{+0.012} & \gain{+0.023} & \loss{-0.025} \\
Qwen3.5-27B & \textbf{\gain{+0.071}} & \gain{+0.040} & \textbf{\gain{+0.275}} & \gain{+0.033} & \neutral{+0.000} & \gain{+0.061} & \textbf{\gain{+0.083}} & \gain{+0.051} \\
Qwen3.5-35B-A3B & \gain{+0.033} & \gain{+0.020} & \gain{+0.050} & \gain{+0.050} & \gain{+0.017} & \gain{+0.036} & \gain{+0.045} & \gain{+0.025} \\
Qwen3.5-9B & \loss{-0.005} & \loss{-0.020} & \neutral{+0.000} & \loss{-0.017} & \gain{+0.017} & \loss{-0.006} & \gain{+0.008} & \loss{-0.006} \\
GLM-4.5-Air & \gain{+0.038} & \textbf{\gain{+0.140}} & \gain{+0.025} & \gain{+0.017} & \loss{-0.017} & \gain{+0.018} & \gain{+0.030} & \gain{+0.006} \\
\addlinespace[2pt]

\rowcolor{gray!12}
\multicolumn{9}{c}{\textbf{Codex}} \\
GPT-5.5 & \textbf{\gain{+0.024}} & \gain{+0.020} & \gain{+0.075} & \gain{+0.050} & \textbf{\loss{-0.033}} & \gain{+0.012} & \gain{+0.015} & \textbf{\gain{+0.025}} \\
GPT-5.4 & \gain{+0.024} & \gain{+0.040} & \gain{+0.075} & \neutral{+0.000} & \neutral{+0.000} & \gain{+0.030} & \gain{+0.030} & \gain{+0.013} \\
DeepSeek-V4-Pro & \gain{+0.024} & \gain{+0.060} & \gain{+0.025} & \textbf{\loss{-0.017}} & \textbf{\gain{+0.033}} & \gain{+0.030} & \loss{-0.008} & \gain{+0.019} \\
Gemma4-31B & \gain{+0.014} & \gain{+0.060} & \neutral{+0.000} & \neutral{+0.000} & \neutral{+0.000} & \loss{-0.012} & \gain{+0.008} & \gain{+0.006} \\
Qwen3.6-27B & \textbf{\loss{-0.038}} & \textbf{\loss{-0.120}} & \textbf{\loss{-0.175}} & \textbf{\gain{+0.083}} & \neutral{+0.000} & \textbf{\loss{-0.030}} & \gain{+0.008} & \textbf{\loss{-0.013}} \\
Qwen3.6-35B-A3B & \gain{+0.014} & \loss{-0.100} & \gain{+0.100} & \gain{+0.033} & \gain{+0.033} & \textbf{\gain{+0.048}} & \textbf{\loss{-0.015}} & \gain{+0.013} \\
Qwen3.5-122B-A10B & \loss{-0.010} & \gain{+0.020} & \loss{-0.075} & \gain{+0.017} & \loss{-0.017} & \loss{-0.006} & \neutral{+0.000} & \loss{-0.013} \\
Qwen3.5-27B & \gain{+0.024} & \neutral{+0.000} & \gain{+0.100} & \gain{+0.017} & \neutral{+0.000} & \gain{+0.012} & \gain{+0.015} & \gain{+0.025} \\
Qwen3.5-35B-A3B & \gain{+0.019} & \gain{+0.020} & \textbf{\gain{+0.125}} & \loss{-0.017} & \loss{-0.017} & \gain{+0.006} & \textbf{\gain{+0.038}} & \neutral{+0.000} \\
Qwen3.5-9B & \gain{+0.024} & \textbf{\gain{+0.100}} & \loss{-0.025} & \gain{+0.017} & \neutral{+0.000} & \neutral{+0.000} & \gain{+0.015} & \gain{+0.025} \\
\bottomrule
\end{tabular}
}
\caption{Skill-induced pass-rate changes by model, difficulty level, and operation family. Values are computed as w/ skill minus w/o skill accuracy; green indicates gains and red indicates regressions. Bold marks the largest gain and largest regression within each harness block and column.}
\label{tab:section4_skill_delta_full}
\end{table*}

\section{Verifier Fidelity Evaluation}
\label{app:verifier_fidelity}

We evaluate verifier fidelity through two complementary experiments. The manual audit measures agreement on actual agent outputs, while the mutation stress test examines whether controlled document corruptions are detected.

\noindent\textbf{Manual artifact-level audit.}
We sampled 128 verifier decisions spanning multiple document formats and difficulty levels. A computer science PhD candidate experienced in document operations inspected the instruction, source artifacts, and submitted output for each case. The auditor assigned a pass or fail judgment based on completion of the requested edits, correctness of native structure, preservation of relevant out-of-scope state, and validity of the required output. We then compared these judgments with the deterministic verifier decisions. As shown in Table~\ref{tab:app_verifier_fidelity}, the verifier agrees with the manual judgments in 122 of 128 cases (95.31\%), with three false passes and three false fails.

One representative false pass correctly reordered the requested slides but also altered the presentation's global canvas dimensions, which were not covered by the verifier. One false fail correctly removed an initial blank page but was rejected by an over-specific predicate tied to a particular low-level page-break representation. We consider these cases acceptable residual edge cases for deterministic artifact validation, as they involve peripheral document properties or implementation-level representations and occur infrequently enough to avoid materially affecting the benchmark conclusions.

\noindent\textbf{Mutation-based stress test.}
We selected 36 outputs accepted by both the manual audit and the verifier, then introduced five controlled mutations into each output. These mutations altered requested content, corrupted native structure, violated preservation requirements, deleted the required output, or damaged the file while retaining its extension. The verifier detects 174 of the 180 mutations (96.67\%), including all missing-output and parse-corruption cases.

To further analyze the undetected mutation cases, we conducted a case study for error analysis. One representative acceptable mistake is an Excel editing task where the required operation was a localized rewrite, while the mutation only affected a worksheet-view detail unrelated to the task. Since all task-related content and the workbook itself remained valid, and the required operation did not involve that worksheet view totally, we consider this case acceptable under the task scope. Overall, these results support the reliability of the verifier suite, while the few missed cases help quantify concrete boundary cases.

\begin{table}[t]
\centering
\scriptsize
\setlength{\tabcolsep}{5pt}
\renewcommand{\arraystretch}{1.08}
\begin{tabular}{@{}lr@{}}
\toprule
\textbf{Evaluation outcome} & \textbf{Result} \\
\midrule
\multicolumn{2}{@{}l}{\textit{Manual audit (128 decisions)}} \\
Agreement & 122/128 (95.31\%) \\
False pass & 3 \\
False fail & 3 \\
\midrule
\multicolumn{2}{@{}l}{\textit{Mutation stress test (36 tasks)}} \\
Requested-content error & 35/36 (97.22\%) \\
Native-structure error & 33/36 (91.67\%) \\
Preservation error & 34/36 (94.44\%) \\
Required output deleted & 36/36 (100.00\%) \\
Same-extension parse corruption & 36/36 (100.00\%) \\
\textbf{Overall} & \textbf{174/180 (96.67\%)} \\
\bottomrule
\end{tabular}
\caption{Verifier-fidelity results. False passes and false fails are defined relative to manual artifact-level judgments. Mutation rates report the proportion of injected violations detected by the verifier.}
\label{tab:app_verifier_fidelity}
\end{table}

\section{Representative Verifier Snippets}
\label{sec:app_verifier_snippets}

Each task verifier is written as a deterministic post-condition checker over the submitted artifact. The goal is not to enforce a single edit script, but to check whether the final document satisfies the semantic and structural contract of the task. Figures~\ref{fig:app_excel_verifier}--\ref{fig:app_pdf_verifier} show representative snippets from four formats.

\paragraph{Excel formula verification.}
Spreadsheet tasks often admit visually plausible but semantically wrong outputs: a model can type the correct-looking number while failing to preserve executable spreadsheet logic. The Excel verifier therefore checks both \emph{formula presence} and \emph{formula semantics}. It accepts several valid formula families, but rejects static values and formulas that are not grounded in the required source ranges.

\begin{figure}[t]
\centering
\setlength{\belowcaptionskip}{-0.4cm}
\setlength{\fboxsep}{6pt}
\fcolorbox{black!30}{gray!4}{\begin{minipage}{0.92\columnwidth}
{\ttfamily\scriptsize
def verify\_task():\par
\quad ws = load\_xlsx\_pair(OUTPUT\_PATH)[0]["Cost Tracker"]\par
\quad assert ws["I3"].value.startswith("=")\par
\quad assert ws["I4"].value.startswith("=")\par
\quad formula = normalize\_text(ws["I3"].value)\par
\quad assert "sumif" in formula or "sumproduct" in formula\par
\quad mapping = code\_map(wb["Code Map"])\par
\quad assert\_formula\_total(ws, mapping, "I3", "Maintenance", 9125.00)\par
\quad assert\_formula\_total(ws, mapping, "I4", "Operations", 3385.00)\par
}
\end{minipage}}
\caption{Representative Excel verifier snippet. The check distinguishes executable formulas from static numeric answers and recomputes expected totals from the workbook state.}
\label{fig:app_excel_verifier}
\end{figure}

\paragraph{Word hierarchy and table verification.}
Word tasks frequently involve hidden document structure: heading styles, section boundaries, and native tables are not equivalent to plain visible text. The Word verifier reads the \texttt{.docx} object model, extracts styled heading paragraphs, and inspects native table cells directly. This prevents an agent from passing by only drawing text that looks like a heading or table.

\begin{figure}[t]
\centering
\setlength{\fboxsep}{6pt}
\fcolorbox{black!30}{gray!4}{\begin{minipage}{0.92\columnwidth}
{\ttfamily\scriptsize
def verify\_task():\par
\quad doc = Document(OUTPUT\_PATH)\par
\quad headings = heading\_paragraphs(doc)\par
\quad assert headings == EXPECT["heading\_sequence"]\par
\quad for spec in EXPECT["paragraphs\_under\_headings"]:\par
\quad\quad actual = paragraphs\_under\_heading(doc, spec["heading"])\par
\quad\quad assert actual == spec["exact\_lines"]\par
\quad table = doc.tables[EXPECT["table"]["index"]]\par
\quad assert table.cell(0, 0).text.strip() == "Item"\par
\quad assert len(table.columns) == EXPECT["table"]["cols"]\par
}
\end{minipage}}
\caption{Representative Word verifier snippet. The verifier grades heading hierarchy and native table state through the Word object model.}
\label{fig:app_word_verifier}
\end{figure}

\paragraph{PowerPoint native table verification.}
PowerPoint is especially prone to destructive visual patches: an agent may overlay text boxes that resemble a table while leaving the underlying slide table unchanged. The verifier therefore searches for a real \texttt{shape.has\_table} object and validates the native table grid and preserved row content.

\begin{figure}[t]
\centering
\setlength{\belowcaptionskip}{-0.4cm}
\setlength{\fboxsep}{6pt}
\fcolorbox{black!30}{gray!4}{\begin{minipage}{0.92\columnwidth}
{\ttfamily\scriptsize
def verify\_task():\par
\quad prs = Presentation(OUTPUT\_PATH)\par
\quad slide = prs.slides[0]\par
\quad table\_shape = next(s for s in slide.shapes if s.has\_table)\par
\quad table = table\_shape.table\par
\quad headers = [table.cell(0, c).text.strip()\par
\quad\quad for c in range(len(table.columns))]\par
\quad assert headers == ["Milestone", "Owner", "Due Date", "Status"]\par
\quad row = [table.cell(1, c).text.strip() for c in range(len(table.columns))]\par
\quad assert row == ["Variation order approved", "", "22 Apr", "Open"]\par
}
\end{minipage}}
\caption{Representative PowerPoint verifier snippet. The check requires a native PowerPoint table rather than visually similar text boxes.}
\label{fig:app_ppt_verifier}
\end{figure}

\paragraph{PDF page-order and outline verification.}
PDF tasks combine visual page content with document navigation state. The PDF verifier therefore checks the visible page order and the outline tree together: page text confirms that the physical sequence is correct, while extracted outline titles and levels confirm that bookmarks were rebuilt rather than merely ignored.

\begin{figure}[t]
\centering
\setlength{\belowcaptionskip}{-0.4cm}
\setlength{\fboxsep}{6pt}
\fcolorbox{black!30}{gray!4}{\begin{minipage}{0.92\columnwidth}
{\ttfamily\scriptsize
def verify\_task():\par
\quad pages = pdf\_page\_texts(OUTPUT\_PATH)\par
\quad assert normalize\_text(pages[0]).startswith("scope")\par
\quad assert normalize\_text(pages[1]).startswith("findings")\par
\quad title = normalize\_text(pages[2])\par
\quad assert title.startswith("recommendations")\par
\quad outline = extract\_pdf\_outline\_titles(OUTPUT\_PATH)\par
\quad expected = [(0,"Scope"), (0,"Findings"), (1,"Site A"),\par
\quad\quad (1,"Site B"), (0,"Recommendations"), (0,"Appendix")]\par
\quad assert outline == expected\par
}
\end{minipage}}
\caption{Representative PDF verifier snippet. The verifier jointly checks page order and bookmark hierarchy.}
\label{fig:app_pdf_verifier}
\end{figure}

\section{Model Serving Details}
\label{sec:app_model_serving}

All evaluated models were publicly accessible at the time of evaluation. Their access and serving configurations are summarized as follows:

\begin{table*}[t]
\centering
\small
\setlength{\belowcaptionskip}{-0.4cm}
\setlength{\tabcolsep}{5pt}
\begin{tabular}{@{}lllll@{}}
\toprule
\textbf{Model} & \textbf{Served name} & \textbf{Max model lens} & \textbf{TP} &
\textbf{Parser} \\
\midrule
Qwen3.6-27B & \texttt{local-qwen36-27b} & 131072 & 8 & \texttt{qwen3\_coder} \\
Qwen3.6-35B-A3B & \texttt{local-qwen36-35b} & 131072 & 8 & \texttt{qwen3\_coder} \\
Qwen3.5-122B-A10B & \texttt{local-qwen35-122b} & 262144 & 8 & \texttt{qwen3\_coder} \\
Qwen3.5-27B & \texttt{local-qwen35-27b} & 131072 & 8 & \texttt{qwen3\_coder} \\
Qwen3.5-35B-A3B & \texttt{local-qwen35-35b} & 131072 & 8 & \texttt{qwen3\_coder} \\
Qwen3.5-9B & \texttt{local-qwen35-9b} & 131072 & 8 & \texttt{qwen3\_coder} \\
Gemma4-31B & \texttt{local-gemma4-31b} & 131072 & 8 & \texttt{gemma4} \\
GLM-4.5-Air & \texttt{local-glm45-air} & 131072 & 8 & \texttt{hermes} \\
\bottomrule
\end{tabular}
\caption{Model serving configurations for locally evaluated open-weight models.}
\label{tab:app_model_serving}
\end{table*}

\noindent$\bullet$ \textbf{API access.} GPT-5.5 and GPT-5.4 were accessed through the OpenAI API. Claude Sonnet 4.6 was accessed through the Anthropic API. DeepSeek-V4-Pro was accessed through the DeepSeek API.

\noindent$\bullet$ \textbf{Local deployment.} The Qwen, Gemma, and GLM models were served locally through vLLM~\citep{kwon2023efficient} on a single A800 node with 8 NVIDIA A800-SXM4-80GB GPUs, CUDA 12.8, and vLLM 0.19.1. Models were loaded in \texttt{torch.bfloat16} with tensor parallel size 8, matching the 8 visible A800 GPUs. GPU memory utilization was set to 0.90 by default, while maximum context length was selected per model as listed in Table~\ref{tab:app_model_serving}. vLLM automatic tool choice was enabled, with a model family specific parser used to normalize tool calls before they reached the harness. Claude Code runs used the Anthropic compatible endpoint exposed by vLLM, while Codex, Terminus-2, and DocTools used its OpenAI-compatible endpoint. Qwen3.6 and Gemma4 Codex runs additionally used explicit chat templates. Server-side decoding followed the harness request parameters, while the serving process fixed model loading, context length, parallelism, and tool-call parsing.

\section{Harness Configuration Details}
\label{sec:app_harness_config}

All harnesses are executed through the same Harbor task lifecycle. For each task, Harbor builds or reuses the task Docker image, mounts the task workspace, launches the selected agent inside an isolated environment, collects the submitted artifact, and then runs the deterministic verifier. This design keeps the task files, agent runtime, and verifier separated: the agent edits only the workspace it receives, while the verifier evaluates the final artifact after the agent exits.

\noindent\textbf{Docker task images.}
Every task contains an \texttt{environment/Dockerfile}. The Dockerfile starts from a harness-specific base image, copies the source documents into the container, and installs task-local document skills when required. Benchmark scripts preflight the base image before execution, ensuring that a configuration fails early if the expected CLI or image is missing.

\noindent$\bullet$ \textbf{Codex.}
Codex runs use dedicated container images with the corresponding harness versions and required dependencies preinstalled for frontier-model and open-weight configurations. The harness provides an open shell, file-system access, and script execution. Skill and no-skill settings differ only in whether document skills are materialized into the task image and exposed to the agent.

\noindent$\bullet$ \textbf{Claude Code.}
Claude Code runs use dedicated container images with the corresponding harness version and dependencies preinstalled. The harness provides an interactive coding-agent loop with terminal execution and file-system feedback. For vLLM-backed runs, adaptive thinking is disabled so that open-weight models operate through the same external tool loop without a provider-specific reasoning budget.

\noindent$\bullet$ \textbf{Terminus-2.}
Terminus-2 exposes a stateful shell with full file-system feedback. We use a 200-turn limit, a JSON action parser, and temperature 0.2. Terminus-2 does not receive document-specific skills, it measures what a compact shell agent can accomplish through general programming and file inspection alone.

\noindent$\bullet$ \textbf{DocTools.}
DocTools runs use a dedicated container image with the corresponding harness implementation and dependencies preinstalled. The harness exposes structured document-editing tools rather than an unrestricted shell. We use a temperature of 0.0 and a small fixed step budget, allowing us to evaluate a document-specific tool interface separately from the broader flexibility of programmable shell environments.

\paragraph{Skill and no-skill materialization.}
Skill-enabled tasks copy the same released document skills into the task image and expose them through task metadata. No-skill runs are produced by a deterministic materialization script: it copies the benchmark project, removes skill directory declarations, strips skill advertisements from task instructions, removes Dockerfile skill-copy steps, and deletes task-local skill folders. As a result, w/ skill and w/o skill settings share the same source documents, verifiers, task structure, and harness interface, procedural document skills are the only controlled difference.

\clearpage
\onecolumn

{
\scriptsize
\setlength{\LTleft}{0pt}
\setlength{\LTright}{0pt}
\setlength{\tabcolsep}{2pt}
\renewcommand{\arraystretch}{1.05}

\begin{longtable}{@{}p{2.15cm}p{0.85cm}p{8.15cm}p{2.55cm}p{1.35cm}@{}}

\caption{Full DocOps 210 task list. Input and output types are normalized file-format labels for the source artifacts and required submission.}
\label{tab:app_full_task_list}\\

\toprule
\textbf{Task ID} & \textbf{Level} & \textbf{Operation labels} & \textbf{Input types} & \textbf{Output type} \\
\midrule
\endfirsthead

\toprule
\textbf{Task ID} & \textbf{Level} & \textbf{Operation labels} & \textbf{Input types} & \textbf{Output type} \\
\midrule
\endhead

\bottomrule
\endfoot

\bottomrule
\endlastfoot

excel\_001 & L1 & C4 Computation & Excel & Excel \\
excel\_002 & L1 & C4 Computation & Excel & Excel \\
excel\_003 & L1 & F2 Highlighting & Excel & Excel \\
excel\_004 & L1 & S4 Table/Sheet operations & Excel & Excel \\
excel\_005 & L1 & C5 Reasoning & Excel & TXT \\
excel\_006 & L1 & C3 Generation & Excel & Excel \\
excel\_007 & L1 & C1 Extraction & Excel & TXT \\
excel\_008 & L1 & C2 Editing & Excel & Excel \\
excel\_009 & L1 & F1 Style consistency & Excel & Excel \\
excel\_010 & L1 & F3 Layout control & Excel & Excel \\
excel\_011 & L1 & F4 Theme transfer & Excel & Excel \\
excel\_012 & L1 & S1 Insert/Delete & Excel & Excel \\
excel\_013 & L1 & S2 Reorder & Excel & Excel \\
excel\_014 & L1 & S3 Hierarchy editing & Excel & Excel \\
pdf\_001 & L1 & C1 Extraction & PDF & TXT \\
pdf\_002 & L1 & C5 Reasoning & PDF & TXT \\
pdf\_003 & L1 & F2 Highlighting & PDF & PDF \\
pdf\_004 & L1 & S2 Reorder & PDF & PDF \\
pdf\_005 & L1 & C2 Editing & PDF & PDF \\
pdf\_006 & L1 & C3 Generation & PDF & PDF \\
pdf\_007 & L1 & F1 Style consistency & PDF & PDF \\
pdf\_008 & L1 & F3 Layout control & PDF & PDF \\
pdf\_009 & L1 & F4 Theme transfer & PDF & PDF \\
pdf\_010 & L1 & S1 Insert/Delete & PDF & PDF \\
pdf\_011 & L1 & S3 Hierarchy editing & PDF & PDF \\
pdf\_012 & L1 & S4 Table/Sheet operations & PDF & PDF \\
ppt\_001 & L1 & C3 Generation & PPT & PPT \\
ppt\_002 & L1 & F3 Layout control & PPT & PPT \\
ppt\_003 & L1 & F4 Theme transfer & PPT & PPT \\
ppt\_004 & L1 & S2 Reorder & PPT & PPT \\
ppt\_005 & L1 & S4 Table/Sheet operations & PPT & PPT \\
ppt\_006 & L1 & S1 Insert/Delete & PPT & PPT \\
ppt\_007 & L1 & C1 Extraction & PPT & TXT \\
ppt\_008 & L1 & C2 Editing & PPT & PPT \\
ppt\_009 & L1 & C5 Reasoning & PPT & TXT \\
ppt\_010 & L1 & F1 Style consistency & PPT & PPT \\
ppt\_011 & L1 & F2 Highlighting & PPT & PPT \\
ppt\_012 & L1 & S3 Hierarchy editing & PPT & PPT \\
word\_001 & L1 & S3 Hierarchy editing & Word & Word \\
word\_002 & L1 & S1 Insert/Delete & Word & Word \\
word\_003 & L1 & C2 Editing & Word & Word \\
word\_004 & L1 & F1 Style consistency & Word & Word \\
word\_005 & L1 & C1 Extraction & Word & TXT \\
word\_006 & L1 & S2 Reorder & Word & Word \\
word\_007 & L1 & C3 Generation & Word & Word \\
word\_008 & L1 & C5 Reasoning & Word & TXT \\
word\_009 & L1 & F2 Highlighting & Word & Word \\
word\_010 & L1 & F3 Layout control & Word & Word \\
word\_011 & L1 & F4 Theme transfer & Word & Word \\
word\_012 & L1 & S4 Table/Sheet operations & Word & Word \\
excelc\_001 & L2 & C5 Reasoning; C2 Editing; F2 Highlighting & Excel & Excel \\
excelc\_002 & L2 & S4 Table/Sheet operations; C4 Computation; F4 Theme transfer & Excel & Excel \\
excelc\_003 & L2 & S2 Reorder; C3 Generation & Excel & Excel \\
excelc\_004 & L2 & S3 Hierarchy editing; F3 Layout control & Excel & Excel \\
excelc\_005 & L2 & C1 Extraction; F2 Highlighting & Excel & Excel \\
excelc\_006 & L2 & C2 Editing; F1 Style consistency & Excel & Excel \\
excelc\_007 & L2 & S1 Insert/Delete; C3 Generation & Excel & Excel \\
excelc\_008 & L2 & S2 Reorder; F4 Theme transfer & Excel & Excel \\
excelc\_009 & L2 & C1 Extraction; S4 Table/Sheet operations & Excel & Excel \\
excelc\_010 & L2 & C4 Computation; F2 Highlighting & Excel & Excel \\
pdfc\_001 & L2 & C5 Reasoning; F2 Highlighting & PDF & PDF \\
pdfc\_002 & L2 & S2 Reorder; S3 Hierarchy editing & PDF & PDF \\
pdfc\_003 & L2 & C3 Generation; F4 Theme transfer & PDF & PDF \\
pdfc\_004 & L2 & S4 Table/Sheet operations; F2 Highlighting & PDF & PDF \\
pdfc\_005 & L2 & C1 Extraction; C3 Generation & PDF & PDF \\
pdfc\_006 & L2 & C2 Editing; F1 Style consistency & PDF & PDF \\
pdfc\_007 & L2 & S1 Insert/Delete; F4 Theme transfer & PDF & PDF \\
pdfc\_008 & L2 & S2 Reorder; C3 Generation & PDF & PDF \\
pdfc\_009 & L2 & C5 Reasoning; C2 Editing & PDF & PDF \\
pdfc\_010 & L2 & F4 Theme transfer; F3 Layout control & PDF & PDF \\
pptc\_001 & L2 & F4 Theme transfer; F1 Style consistency; F3 Layout control & PPT & PPT \\
pptc\_002 & L2 & S2 Reorder; C3 Generation & PPT & PPT \\
pptc\_003 & L2 & C5 Reasoning; F2 Highlighting & PPT & PPT \\
pptc\_004 & L2 & S4 Table/Sheet operations; C2 Editing & PPT & PPT \\
pptc\_005 & L2 & C1 Extraction; F2 Highlighting & PPT & PPT \\
pptc\_006 & L2 & C2 Editing; F1 Style consistency & PPT & PPT \\
pptc\_007 & L2 & S1 Insert/Delete; C3 Generation & PPT & PPT \\
pptc\_008 & L2 & S3 Hierarchy editing; F3 Layout control & PPT & PPT \\
pptc\_009 & L2 & S2 Reorder; F4 Theme transfer & PPT & PPT \\
pptc\_010 & L2 & C1 Extraction; S4 Table/Sheet operations & PPT & PPT \\
wordc\_001 & L2 & S3 Hierarchy editing; F1 Style consistency & Word & Word \\
wordc\_002 & L2 & S2 Reorder; C3 Generation & Word & Word \\
wordc\_003 & L2 & C5 Reasoning; F2 Highlighting & Word & Word \\
wordc\_004 & L2 & S4 Table/Sheet operations; C2 Editing & Word & Word \\
wordc\_005 & L2 & C1 Extraction; F2 Highlighting & Word & Word \\
wordc\_006 & L2 & C2 Editing; F1 Style consistency & Word & Word \\
wordc\_007 & L2 & S1 Insert/Delete; C3 Generation & Word & Word \\
wordc\_008 & L2 & S2 Reorder; F4 Theme transfer & Word & Word \\
wordc\_009 & L2 & C1 Extraction; S4 Table/Sheet operations & Word & Word \\
wordc\_010 & L2 & C5 Reasoning; C2 Editing & Word & Word \\
excelwf\_001 & L3 & S4 Table/Sheet operations; C4 Computation; S2 Reorder; F2 Highlighting; F4 Theme transfer & Excel & Excel \\
excelwf\_002 & L3 & S2 Reorder; F3 Layout control; C4 Computation; F2 Highlighting; S1 Insert/Delete & Excel & Excel \\
excelwf\_003 & L3 & C4 Computation; S1 Insert/Delete; S2 Reorder; F2 Highlighting; F4 Theme transfer & Excel & Excel \\
excelwf\_004 & L3 & S4 Table/Sheet operations; C4 Computation; F1 Style consistency; S2 Reorder & Excel & Excel \\
excelwf\_005 & L3 & S1 Insert/Delete; S2 Reorder; C4 Computation; F4 Theme transfer; F1 Style consistency & Excel & Excel \\
pdfwf\_001 & L3 & S2 Reorder; S1 Insert/Delete; C3 Generation; S3 Hierarchy editing & PDF & PDF \\
pdfwf\_002 & L3 & F4 Theme transfer; C3 Generation; F3 Layout control; S2 Reorder & PDF & PDF \\
pdfwf\_003 & L3 & S1 Insert/Delete; S2 Reorder; S3 Hierarchy editing; C3 Generation & PDF & PDF \\
pdfwf\_004 & L3 & S2 Reorder; C3 Generation; S3 Hierarchy editing; F4 Theme transfer & PDF & PDF \\
pdfwf\_005 & L3 & S2 Reorder; S3 Hierarchy editing; F4 Theme transfer; C3 Generation & PDF & PDF \\
pptwf\_001 & L3 & S2 Reorder; S1 Insert/Delete; C3 Generation; F1 Style consistency; F4 Theme transfer & PPT & PPT \\
pptwf\_002 & L3 & F4 Theme transfer; F1 Style consistency; S3 Hierarchy editing; C2 Editing & PPT & PPT \\
pptwf\_003 & L3 & S1 Insert/Delete; S4 Table/Sheet operations; F3 Layout control; F4 Theme transfer & PPT & PPT \\
pptwf\_004 & L3 & C2 Editing; S2 Reorder; F3 Layout control; F1 Style consistency & PPT & PPT \\
pptwf\_005 & L3 & C2 Editing; S2 Reorder; F1 Style consistency; F3 Layout control & PPT & PPT \\
wordwf\_001 & L3 & C2 Editing; S2 Reorder; S3 Hierarchy editing; F1 Style consistency; S1 Insert/Delete & Word & Word \\
wordwf\_002 & L3 & S2 Reorder; S3 Hierarchy editing; F3 Layout control; F1 Style consistency & Word & Word \\
wordwf\_003 & L3 & C2 Editing; F1 Style consistency; S4 Table/Sheet operations & Word & Word \\
wordwf\_004 & L3 & S3 Hierarchy editing; S4 Table/Sheet operations; C3 Generation; C5 Reasoning; F1 Style consistency & Word & Word \\
wordwf\_005 & L3 & C2 Editing; C3 Generation; F1 Style consistency; F3 Layout control; S1 Insert/Delete; S3 Hierarchy editing; S4 Table/Sheet operations & Word & Word \\
excelxr\_001 & L4 & C1 Extraction; C5 Reasoning; S2 Reorder; S4 Table/Sheet operations & Excel, PDF, Word & Excel \\
excelxr\_002 & L4 & C1 Extraction; C4 Computation; F2 Highlighting; S4 Table/Sheet operations & Excel, PDF, Word & Excel \\
excelxr\_003 & L4 & C1 Extraction; C5 Reasoning; S4 Table/Sheet operations & Excel, PDF, Word & Excel \\
excelxr\_004 & L4 & C1 Extraction; C2 Editing; S4 Table/Sheet operations & Excel, PDF, Word & Excel \\
excelxr\_005 & L4 & C1 Extraction; S3 Hierarchy editing; S4 Table/Sheet operations & Excel, PDF, Word & Excel \\
pdfxr\_001 & L4 & C1 Extraction; C3 Generation; S1 Insert/Delete; S2 Reorder; S3 Hierarchy editing & PDF, Word & PDF \\
pdfxr\_002 & L4 & C1 Extraction; C2 Editing; S1 Insert/Delete; S2 Reorder; S3 Hierarchy editing & Excel, PDF, Word, PNG & PDF \\
pdfxr\_003 & L4 & C1 Extraction; C5 Reasoning; S1 Insert/Delete; S2 Reorder; S3 Hierarchy editing & Excel, PDF, Word, PNG & PDF \\
pdfxr\_004 & L4 & C1 Extraction; C2 Editing; S1 Insert/Delete; S2 Reorder; S3 Hierarchy editing & Excel, PDF, Word, PNG & PDF \\
pdfxr\_005 & L4 & C1 Extraction; C5 Reasoning; S2 Reorder; S3 Hierarchy editing & Excel, PDF, Word, PNG & PDF \\
pptxr\_001 & L4 & C1 Extraction; C4 Computation; F1 Style consistency; S1 Insert/Delete; S4 Table/Sheet operations & Excel, PPT, Word, PNG & PPT \\
pptxr\_002 & L4 & C1 Extraction; C2 Editing; F1 Style consistency; S1 Insert/Delete; S2 Reorder; S4 Table/Sheet operations & Excel, PPT, Word, PNG & PPT \\
pptxr\_003 & L4 & C1 Extraction; C2 Editing; C3 Generation; F1 Style consistency; S1 Insert/Delete; S4 Table/Sheet operations & Excel, PPT, PDF, Word, PNG & PPT \\
pptxr\_004 & L4 & C1 Extraction; C4 Computation; F1 Style consistency; S1 Insert/Delete; S4 Table/Sheet operations & Excel, PPT, Word, PNG & PPT \\
pptxr\_005 & L4 & C1 Extraction; C2 Editing; C3 Generation; F1 Style consistency; F4 Theme transfer; S1 Insert/Delete & Excel, PPT, PDF, Word & PPT \\
wordxr\_001 & L4 & C1 Extraction; C2 Editing; F1 Style consistency; S3 Hierarchy editing; S4 Table/Sheet operations & Excel, PDF, Word, PNG & Word \\
wordxr\_002 & L4 & C1 Extraction; C2 Editing; C5 Reasoning; F1 Style consistency; S4 Table/Sheet operations & Excel, PDF, Word & Word \\
wordxr\_003 & L4 & C1 Extraction; C2 Editing; F1 Style consistency; F2 Highlighting; S3 Hierarchy editing; S4 Table/Sheet operations & Excel, PDF, Word, PNG & Word \\
wordxr\_004 & L4 & C1 Extraction; C2 Editing; F1 Style consistency; S4 Table/Sheet operations & Excel, PDF, Word & Word \\
wordxr\_005 & L4 & C1 Extraction; C2 Editing; F1 Style consistency; F2 Highlighting; S3 Hierarchy editing; S4 Table/Sheet operations & Excel, PDF, Word, PNG & Word \\
L3\_v2\_011 & L3 & C1 Extraction; C2 Editing; C3 Generation; C4 Computation; C5 Reasoning; F1 Style consistency; F2 Highlighting; F3 Layout control; S1 Insert/Delete; S2 Reorder; S3 Hierarchy editing; S4 Table/Sheet operations & Excel & Excel \\
L3\_v2\_012 & L3 & C2 Editing; S2 Reorder & Excel & Excel \\
L3\_v2\_013 & L3 & C2 Editing; S2 Reorder & Excel & Excel \\
L3\_v2\_014 & L3 & C2 Editing; C4 Computation; S4 Table/Sheet operations & Excel & Excel \\
L3\_v2\_015 & L3 & C2 Editing; C4 Computation; S4 Table/Sheet operations & Excel & Excel \\
L3\_v2\_016 & L3 & C1 Extraction; C2 Editing; C3 Generation; C4 Computation; C5 Reasoning; F1 Style consistency; F2 Highlighting; F3 Layout control; F4 Theme transfer; S1 Insert/Delete; S2 Reorder; S3 Hierarchy editing; S4 Table/Sheet operations & Excel & Excel \\
L3\_v2\_017 & L3 & C2 Editing; S2 Reorder & Excel & Excel \\
L3\_v2\_018 & L3 & C2 Editing; S2 Reorder & Excel & Excel \\
L3\_v2\_019 & L3 & C2 Editing; C4 Computation; S4 Table/Sheet operations & Excel & Excel \\
L3\_v2\_020 & L3 & C1 Extraction; C2 Editing; C3 Generation; C4 Computation; F1 Style consistency; F2 Highlighting; F3 Layout control; F4 Theme transfer; S1 Insert/Delete; S2 Reorder; S3 Hierarchy editing; S4 Table/Sheet operations & Excel & Excel \\
L3\_v2\_031 & L3 & C2 Editing; C4 Computation; F1 Style consistency; S4 Table/Sheet operations & PDF & PDF \\
L3\_v2\_032 & L3 & C2 Editing; S2 Reorder & PDF & PDF \\
L3\_v2\_033 & L3 & C2 Editing; S2 Reorder & PDF & PDF \\
L3\_v2\_034 & L3 & C2 Editing; C4 Computation; F1 Style consistency; S4 Table/Sheet operations & PDF & PDF \\
L3\_v2\_035 & L3 & C2 Editing; S2 Reorder & PDF & PDF \\
L3\_v2\_036 & L3 & C2 Editing; S2 Reorder & PDF & PDF \\
L3\_v2\_037 & L3 & C2 Editing; C4 Computation; F1 Style consistency; S4 Table/Sheet operations & PDF & PDF \\
L3\_v2\_038 & L3 & C2 Editing; C4 Computation; F1 Style consistency; S4 Table/Sheet operations & PDF & PDF \\
L3\_v2\_039 & L3 & C2 Editing; C4 Computation; F1 Style consistency; S4 Table/Sheet operations & PDF & PDF \\
L3\_v2\_040 & L3 & C2 Editing; C4 Computation; F1 Style consistency; S4 Table/Sheet operations & PDF & PDF \\
L3\_v2\_021 & L3 & C2 Editing; C4 Computation; F1 Style consistency; S4 Table/Sheet operations & PPT & PPT \\
L3\_v2\_022 & L3 & C2 Editing; S2 Reorder & PPT & PPT \\
L3\_v2\_023 & L3 & C2 Editing; S2 Reorder & PPT & PPT \\
L3\_v2\_024 & L3 & C2 Editing; S2 Reorder & PPT & PPT \\
L3\_v2\_025 & L3 & C2 Editing; S2 Reorder & PPT & PPT \\
L3\_v2\_026 & L3 & C2 Editing; C4 Computation; F1 Style consistency; S4 Table/Sheet operations & PPT & PPT \\
L3\_v2\_027 & L3 & C2 Editing; C4 Computation; F1 Style consistency; S4 Table/Sheet operations & PPT & PPT \\
L3\_v2\_028 & L3 & C2 Editing; C4 Computation; F1 Style consistency; S4 Table/Sheet operations & PPT & PPT \\
L3\_v2\_029 & L3 & C1 Extraction; C2 Editing; C3 Generation; C4 Computation; C5 Reasoning; F1 Style consistency; F3 Layout control; S1 Insert/Delete; S2 Reorder; S3 Hierarchy editing & PPT & PPT \\
L3\_v2\_030 & L3 & C1 Extraction; C2 Editing; C3 Generation; C4 Computation; F1 Style consistency; F2 Highlighting; F3 Layout control; F4 Theme transfer; S1 Insert/Delete; S2 Reorder; S3 Hierarchy editing; S4 Table/Sheet operations & PPT & PPT \\
L3\_v2\_001 & L3 & C2 Editing; C4 Computation; F1 Style consistency; F3 Layout control; S3 Hierarchy editing; S4 Table/Sheet operations & Word & Word \\
L3\_v2\_002 & L3 & C2 Editing; C4 Computation; F1 Style consistency; F3 Layout control; S3 Hierarchy editing; S4 Table/Sheet operations & Word & Word \\
L3\_v2\_003 & L3 & C2 Editing; C4 Computation; F1 Style consistency; F3 Layout control; S3 Hierarchy editing; S4 Table/Sheet operations & Word & Word \\
L3\_v2\_004 & L3 & C2 Editing; C4 Computation; F1 Style consistency; F3 Layout control; S3 Hierarchy editing; S4 Table/Sheet operations & Word & Word \\
L3\_v2\_005 & L3 & C2 Editing; S2 Reorder; S3 Hierarchy editing; S1 Insert/Delete & Word & Word \\
L3\_v2\_006 & L3 & C2 Editing; S2 Reorder; S3 Hierarchy editing & Word & Word \\
L3\_v2\_007 & L3 & C2 Editing; S2 Reorder; S3 Hierarchy editing & Word & Word \\
L3\_v2\_008 & L3 & C2 Editing; S2 Reorder; S3 Hierarchy editing & Word & Word \\
L3\_v2\_009 & L3 & C1 Extraction; C2 Editing; C3 Generation; C5 Reasoning; F1 Style consistency; F2 Highlighting; F3 Layout control; S1 Insert/Delete; S2 Reorder; S3 Hierarchy editing & Word & Word \\
L3\_v2\_010 & L3 & C1 Extraction; C2 Editing; C3 Generation; C4 Computation; C5 Reasoning; F1 Style consistency; F2 Highlighting; F3 Layout control; F4 Theme transfer; S1 Insert/Delete; S2 Reorder; S3 Hierarchy editing; S4 Table/Sheet operations & Word & Word \\
L4\_v2\_042 & L4 & C1 Extraction; C3 Generation; C4 Computation; C5 Reasoning; F2 Highlighting; S1 Insert/Delete; S3 Hierarchy editing; S4 Table/Sheet operations & Excel, PDF, Word & Excel, PDF \\
L4\_v2\_043 & L4 & C1 Extraction; C2 Editing; C3 Generation; C4 Computation; C5 Reasoning; F1 Style consistency; F2 Highlighting; S1 Insert/Delete; S2 Reorder; S4 Table/Sheet operations & Excel, PPT, PDF, Word & Excel, PPT \\
L4\_v2\_044 & L4 & C1 Extraction; C4 Computation; F2 Highlighting; S1 Insert/Delete; S4 Table/Sheet operations & Excel, PDF, Word & Excel \\
L4\_v2\_080 & L4 & C1 Extraction; C2 Editing; C3 Generation; C5 Reasoning; F1 Style consistency; F2 Highlighting; S2 Reorder; S3 Hierarchy editing; S4 Table/Sheet operations & Excel, PDF, Word & Excel, PDF \\
L4\_v2\_045 & L4 & C1 Extraction; C2 Editing; C3 Generation; C5 Reasoning; F1 Style consistency; F2 Highlighting; S1 Insert/Delete; S3 Hierarchy editing & Excel, PDF, Word & PDF, Word \\
L4\_v2\_046 & L4 & C1 Extraction; C2 Editing; C3 Generation; C5 Reasoning; F1 Style consistency; F2 Highlighting; S1 Insert/Delete; S3 Hierarchy editing & Excel, PDF, Word & PDF, Word \\
L4\_v2\_049 & L4 & C1 Extraction; C2 Editing; C3 Generation; C4 Computation; C5 Reasoning; F2 Highlighting; S3 Hierarchy editing; S4 Table/Sheet operations & Excel, PDF, Word & Excel, PDF, Word \\
L4\_v2\_050 & L4 & C1 Extraction; C2 Editing; C3 Generation; C4 Computation; C5 Reasoning; S1 Insert/Delete; S3 Hierarchy editing; S4 Table/Sheet operations & Excel, PDF, Word & Excel, PDF, Word \\
L4\_v2\_051 & L4 & C1 Extraction; C2 Editing; C3 Generation; C4 Computation; C5 Reasoning; F1 Style consistency; F2 Highlighting; S2 Reorder; S4 Table/Sheet operations & Excel, PDF, Word & Excel, PDF \\
L4\_v2\_052 & L4 & C1 Extraction; C3 Generation; C4 Computation; C5 Reasoning; F2 Highlighting; S1 Insert/Delete; S2 Reorder; S3 Hierarchy editing; S4 Table/Sheet operations & Excel, PDF, Word & Excel, PDF \\
L4\_v2\_053 & L4 & C1 Extraction; C4 Computation; C5 Reasoning; S4 Table/Sheet operations & Excel, PDF, Word & Excel, PDF \\
L4\_v2\_054 & L4 & C1 Extraction; C4 Computation; C5 Reasoning; S4 Table/Sheet operations & Excel, PDF, Word & Excel, PDF \\
L4\_v2\_055 & L4 & C1 Extraction; C2 Editing; S1 Insert/Delete; S2 Reorder; S3 Hierarchy editing & Excel, PDF, Word & PDF \\
L4\_v2\_056 & L4 & C1 Extraction; C2 Editing; C3 Generation; C5 Reasoning; F1 Style consistency; S1 Insert/Delete; S2 Reorder; S3 Hierarchy editing; S4 Table/Sheet operations & Excel, PPT, PDF, Word & PPT, Word \\
L4\_v2\_057 & L4 & C1 Extraction; C2 Editing; C3 Generation; C5 Reasoning; F1 Style consistency; S1 Insert/Delete; S2 Reorder; S3 Hierarchy editing & Excel, PPT, PDF, Word & PPT, PDF \\
L4\_v2\_058 & L4 & C1 Extraction; C3 Generation; C4 Computation; C5 Reasoning; F1 Style consistency; S1 Insert/Delete; S2 Reorder; S4 Table/Sheet operations & Excel, PPT, PDF, Word & Excel, PPT \\
L4\_v2\_059 & L4 & C1 Extraction; C2 Editing; C3 Generation; C4 Computation; C5 Reasoning; F1 Style consistency; S1 Insert/Delete; S2 Reorder; S4 Table/Sheet operations & Excel, PPT, PDF, Word & Excel, PPT \\
L4\_v2\_060 & L4 & C1 Extraction; C2 Editing; C3 Generation; C4 Computation; C5 Reasoning; F1 Style consistency; F2 Highlighting; S1 Insert/Delete; S2 Reorder; S4 Table/Sheet operations & Excel, PPT, PDF, Word & Excel, PPT \\
L4\_v2\_061 & L4 & C1 Extraction; C4 Computation; C5 Reasoning; S4 Table/Sheet operations & Excel, PDF, Word & Excel, PPT \\
L4\_v2\_062 & L4 & C1 Extraction; C4 Computation; C5 Reasoning; S4 Table/Sheet operations & Excel, PPT, PDF, Word & Excel, PPT \\
L4\_v2\_041 & L4 & C1 Extraction; C4 Computation; C5 Reasoning; S4 Table/Sheet operations & Excel, PDF, Word & Excel, PDF, Word \\
L4\_v2\_047 & L4 & C1 Extraction; C4 Computation; C5 Reasoning; S4 Table/Sheet operations & Excel, PDF, Word & Excel, Word \\
L4\_v2\_048 & L4 & C1 Extraction; C4 Computation; C5 Reasoning; S4 Table/Sheet operations & Excel, PDF, Word & Excel, Word \\
L4\_v2\_063 & L4 & C1 Extraction; C2 Editing; C3 Generation; C5 Reasoning; F1 Style consistency; F2 Highlighting; S1 Insert/Delete; S3 Hierarchy editing & Excel, PDF, Word & PDF, Word \\
L4\_v2\_064 & L4 & C1 Extraction; C2 Editing; C3 Generation; C4 Computation; C5 Reasoning; F2 Highlighting; S1 Insert/Delete; S3 Hierarchy editing; S4 Table/Sheet operations & Excel, PDF, Word & PDF, Word \\
L4\_v2\_065 & L4 & C1 Extraction; C2 Editing; C3 Generation; C5 Reasoning; F1 Style consistency; F2 Highlighting; S2 Reorder; S3 Hierarchy editing & Excel, PPT, PDF, Word & PPT, Word \\
L4\_v2\_066 & L4 & C1 Extraction; C2 Editing; C3 Generation; C4 Computation; C5 Reasoning; F1 Style consistency; F2 Highlighting; S3 Hierarchy editing; S4 Table/Sheet operations & Excel, PDF, Word & Excel, Word \\
L4\_v2\_067 & L4 & C1 Extraction; C2 Editing; C3 Generation; C4 Computation; C5 Reasoning; F2 Highlighting; S1 Insert/Delete; S3 Hierarchy editing; S4 Table/Sheet operations & Excel, PDF, Word & Excel, Word \\
L4\_v2\_068 & L4 & C1 Extraction; C2 Editing; C3 Generation; C4 Computation; C5 Reasoning; F2 Highlighting; S3 Hierarchy editing; S4 Table/Sheet operations & Excel, PDF, Word & Excel, Word \\
L4\_v2\_069 & L4 & C1 Extraction; C2 Editing; C3 Generation; C4 Computation; C5 Reasoning; F2 Highlighting; S3 Hierarchy editing; S4 Table/Sheet operations & Excel, PDF, Word & Excel, Word \\
L4\_v2\_070 & L4 & C1 Extraction; C2 Editing; C3 Generation; C4 Computation; C5 Reasoning; F2 Highlighting; S3 Hierarchy editing; S4 Table/Sheet operations & Excel, PDF, Word & Excel, Word \\
L4\_v2\_071 & L4 & C1 Extraction; C2 Editing; C3 Generation; C4 Computation; C5 Reasoning; F2 Highlighting; S3 Hierarchy editing; S4 Table/Sheet operations & Excel, PDF, Word & Excel, Word \\
L4\_v2\_072 & L4 & C1 Extraction; C2 Editing; C3 Generation; C4 Computation; C5 Reasoning; F2 Highlighting; S3 Hierarchy editing; S4 Table/Sheet operations & Excel, PDF, Word & Excel, Word \\
L4\_v2\_073 & L4 & C1 Extraction; C2 Editing; C3 Generation; C4 Computation; C5 Reasoning; F2 Highlighting; S3 Hierarchy editing; S4 Table/Sheet operations & Excel, PDF, Word & Excel, Word \\
L4\_v2\_074 & L4 & C1 Extraction; C4 Computation; C5 Reasoning; S4 Table/Sheet operations & Excel, PDF, Word & Excel, Word \\
L4\_v2\_075 & L4 & C1 Extraction; C4 Computation; C5 Reasoning; S4 Table/Sheet operations & Excel, PDF, Word & Excel, Word \\
L4\_v2\_076 & L4 & C1 Extraction; C4 Computation; C5 Reasoning; S4 Table/Sheet operations & Excel, PDF, Word & Excel, Word \\
L4\_v2\_077 & L4 & C1 Extraction; C4 Computation; C5 Reasoning; S4 Table/Sheet operations & Excel, PDF, Word & Excel, Word \\
L4\_v2\_078 & L4 & C1 Extraction; C4 Computation; C5 Reasoning; S4 Table/Sheet operations & Excel, PDF, Word & Excel, Word \\
L4\_v2\_079 & L4 & C1 Extraction; C2 Editing; C3 Generation; C5 Reasoning; F1 Style consistency; F2 Highlighting; S1 Insert/Delete; S2 Reorder; S3 Hierarchy editing; S4 Table/Sheet operations & Excel, PDF, Word & Excel, Word \\

\end{longtable}
}

\normalsize
\twocolumn

\end{document}